\documentclass[11pt]{article}
\usepackage[final]{acl}

\usepackage{times}
\usepackage{latexsym}

\usepackage[T1]{fontenc}

\usepackage[utf8]{inputenc}

\usepackage{microtype}

\usepackage{inconsolata}

\usepackage{graphicx}

\usepackage{amsmath}

\usepackage{amsfonts}
\usepackage{subcaption}
\usepackage{cleveref}
\usepackage{booktabs}
\usepackage{bbm}
\usepackage{enumitem}
\usepackage{multirow}

\title{Masked by Consensus: Disentangling Privileged Knowledge in LLM Correctness}

\author{
  \textbf{Tomer Ashuach\textsuperscript{1}} \enspace
  \textbf{Shai Gretz\textsuperscript{2}} \enspace
  \textbf{Yoav Katz\textsuperscript{2}} \enspace
  \textbf{Yonatan Belinkov\textsuperscript{1,3}}
  \textbf{Liat Ein-Dor\textsuperscript{2}} \enspace \\
  \\[-0.8em]
  \textsuperscript{1}Technion -- Israel Institute of Technology \enspace
  \textsuperscript{2}IBM Research \\
  \textsuperscript{3}Kempner Institute, Harvard University \\
  \\[-0.8em]
  \texttt{tomerashuach@campus.technion.ac.il, belinkov@technion.ac.il} \\
  \texttt{\{liate,avishaig,katz\}@il.ibm.com}
}

\begin{document}
\maketitle

\begin{abstract}
Humans use introspection to evaluate their understanding through private internal states inaccessible to external observers. We investigate whether large language models possess similar \emph{privileged knowledge} about answer correctness, information unavailable through external observation. We train correctness classifiers on question representations from both a model's own hidden states and external models, testing whether self-representations provide a performance advantage. On standard evaluation, we find no advantage: self-probes perform comparably to peer-model probes. We hypothesize this is due to high inter-model agreement of answer correctness. To isolate genuine privileged knowledge, we evaluate on \emph{disagreement subsets}, where models produce conflicting predictions. Here, we discover domain-specific privileged knowledge: self-representations consistently outperform peer representations in factual knowledge tasks, but show no advantage in math reasoning. We further localize this domain asymmetry across model layers, finding that the factual advantage emerges progressively from early-to-mid layers onward, consistent with model-specific memory retrieval, while math reasoning shows no consistent advantage at any depth.
\end{abstract}

\begin{figure*}[t]
    \centering
    \includegraphics[width=\linewidth]{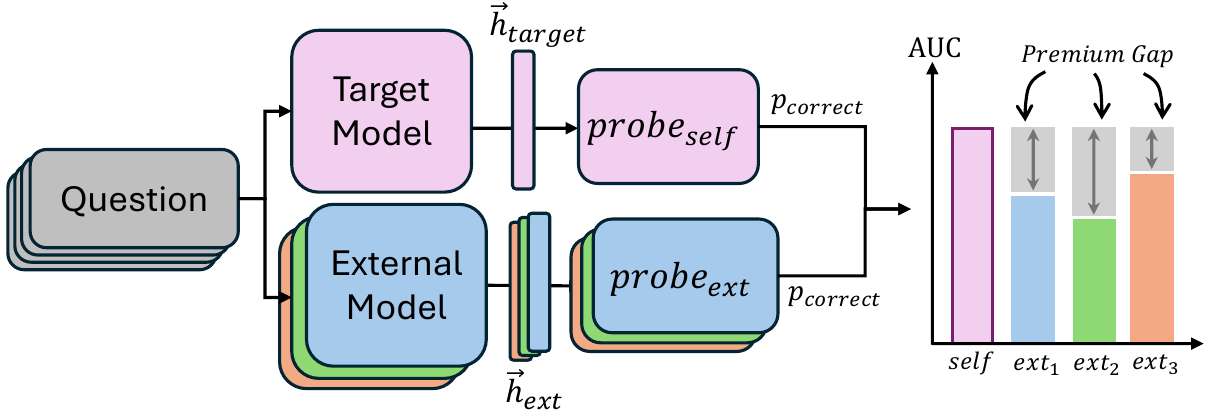}
    \caption{\textbf{Overview of the experimental framework.}
    Questions are input to a target model and to external models, yielding representations $\mathbf{h}_{\text{target}}$ and $\mathbf{h}_{\text{ext}}$. Probes trained on these representations predict answer correctness. We evaluate probe performance using mean AUC averaged over layers and define the \emph{premium gap} as the performance advantage of self over external probes.}
    \label{fig:main_method}
\end{figure*}

\section{Introduction}
In the philosophy of mind, \emph{epistemic privilege} refers to the idea that an agent has special access to its own internal states—information that cannot be fully recovered from external observation alone \citep{alston1971varieties,gertler2010self}.
Inspired by this notion, recent research suggests that large language models (LLMs) encode meta-information about their own outputs, ranging from entity recognition~\citep{ferran2024doknow} and temperature inference~\citep{comsa2025does} to the representation of cognitive-like states~\citep{chen2025imitation,ji2025language}.
A central aspect of this meta-information is \emph{output correctness}: numerous studies have demonstrated that output correctness can be predicted with high accuracy~\citep{kadavath2022language}, primarily via linear probes trained on internal hidden states~\citep{Cencerrado2025No,Seo2025Quantifying}.
This raises a fundamental question: do LLMs have internal correctness signals that are inaccessible to external models? In other words, do they possess \emph{privileged knowledge} about whether their answer will be correct?

Recent findings cast doubt on the existence of privileged knowledge in the context of correctness prediction.
\citet{Chi2025LargeLM} argue that probes primarily detect retrieval activation patterns rather than correctness signals, while \citet{Seo2025Quantifying} and \citet{xiao2025generalized} show that external models can achieve prediction performance comparable to methods that rely on a model's own internal representations, suggesting little to no privileged information exists.
In this paper, we argue that prior conclusions about the absence of privileged knowledge may be premature due to confounded evaluation. Specifically, when external models can exploit proxy signals from shared correctness patterns, genuine privileged knowledge—if it exists—may be masked.
To test whether privileged knowledge exists, we measure the \emph{premium gap}: the performance advantage of a correctness classifier trained on a model's \emph{own} internal representations over one trained on \emph{external} model representations (\Cref{fig:main_method}).
However, this gap may vanish on random samples due to high inter-model agreement.
If models often succeed or fail on the same questions, probes trained on external representations can exploit the external model's own correctness patterns as a \emph{proxy} for the target model's behavior, obscuring whether the target model possesses unique internal signals about its correctness.

To address this challenge, we construct \emph{disagreement subsets}: questions where models produce conflicting correctness labels. By restricting evaluation to these subsets, we eliminate shared correctness patterns and isolate each model's unique behavior, enabling a direct test of whether self-representations contain privileged information unavailable from peer models.

We use this methodology to systematically investigate the existence of privileged knowledge in the context of correctness prediction across five datasets spanning factual knowledge (Mintaka, TriviaQA, HotPotQA) and mathematical reasoning (GSM1K, MATH), using three similar-sized models (Qwen-2.5-7B, Llama-3.1-8B, Gemma-2-9B).

\begin{figure*}[t]
    \centering
    \includegraphics[width=0.98\linewidth]{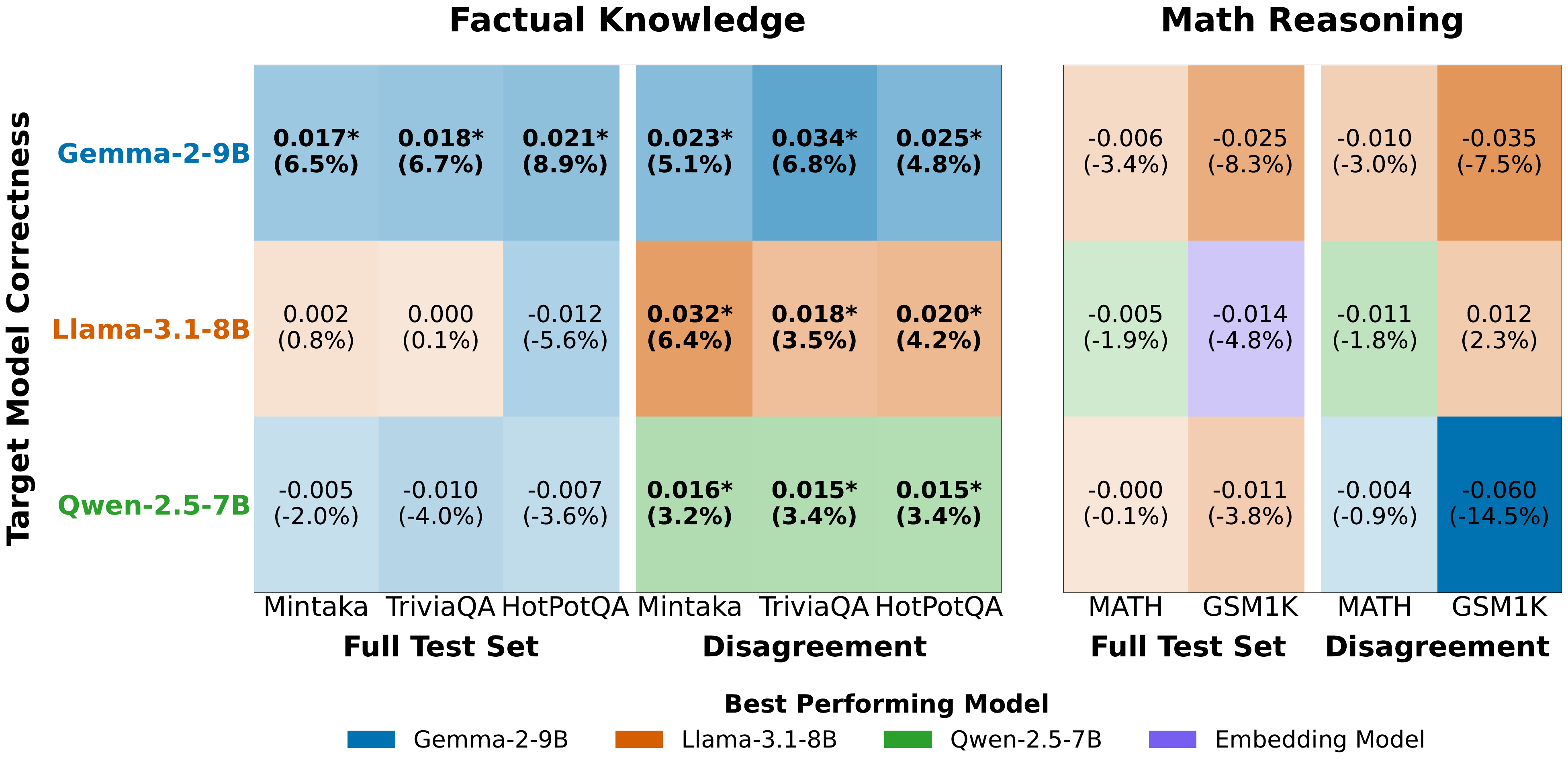}
    \caption{\textbf{Target Model Correctness Prediction.}
    Heatmap of correctness prediction differences across target models, datasets, and test subsets. Each cell reports the AUC difference ($\Delta \mathrm{AUC} = \mathrm{Self} - \mathrm{Best\ External}$), with the percentage of the gap closed shown in parentheses, computed as $\frac{\mathrm{Self}-\mathrm{Best\ External}}{1-\mathrm{Best\ External}}\times 100$. The y-axis lists target models, and cell colors indicate the best-performing source model for each setting. Asterisks (*) denote statistically significant differences (paired $t$-test, $p<0.05$, Bonferroni--Holm correction).}
    \label{fig:heatmap_results}
\end{figure*}

Our results show that measuring the premium gap on a random sample is insufficient to establish privileged knowledge. When a gap does appear for a particular model, that same model also excels at predicting peer correctness—suggesting its stronger representations reflect a general advantage rather than privileged self-knowledge. However, when evaluated on disagreement subsets, a statistically significant premium gap emerges in factual knowledge domains ($\sim$5\%) across all models, providing genuine evidence for privileged knowledge.
In contrast, mathematical reasoning shows no such advantage: probes trained on an external model's representations of the same question perform comparably to those trained on the model's own internal representations, even under disagreement (\Cref{fig:heatmap_results}).
We further localize this domain asymmetry across model layers, finding that the factual advantage emerges progressively from early-to-mid layers onward and strengthens with depth, consistent with model-specific memory retrieval that accumulates through the forward pass, while mathematical reasoning shows no consistent advantage across layers.

We summarize our main findings as follows:
\begin{itemize}[noitemsep, topsep=2pt]
    \item We systematically evaluate correctness prediction across five datasets and three models, demonstrating that the premium gap effectively vanishes when tested against strong external model baselines.
    \item We identify \emph{inter-model agreement} as a critical confounder: probes leverage shared difficulty patterns to predict correctness without needing access to the target's internal state.
    \item We introduce evaluation on \emph{disagreement subsets} to isolate internal signals, revealing that genuine privileged knowledge is domain-specific: it emerges in factual tasks but remains absent in mathematical reasoning.
    \item We localize this domain asymmetry across network depth, showing that the factual advantage emerges progressively from early-to-mid layers onward, while mathematical reasoning shows no consistent advantage at any depth.
\end{itemize}

\begin{figure*}[t]
    \centering  \includegraphics[width=0.9\linewidth]{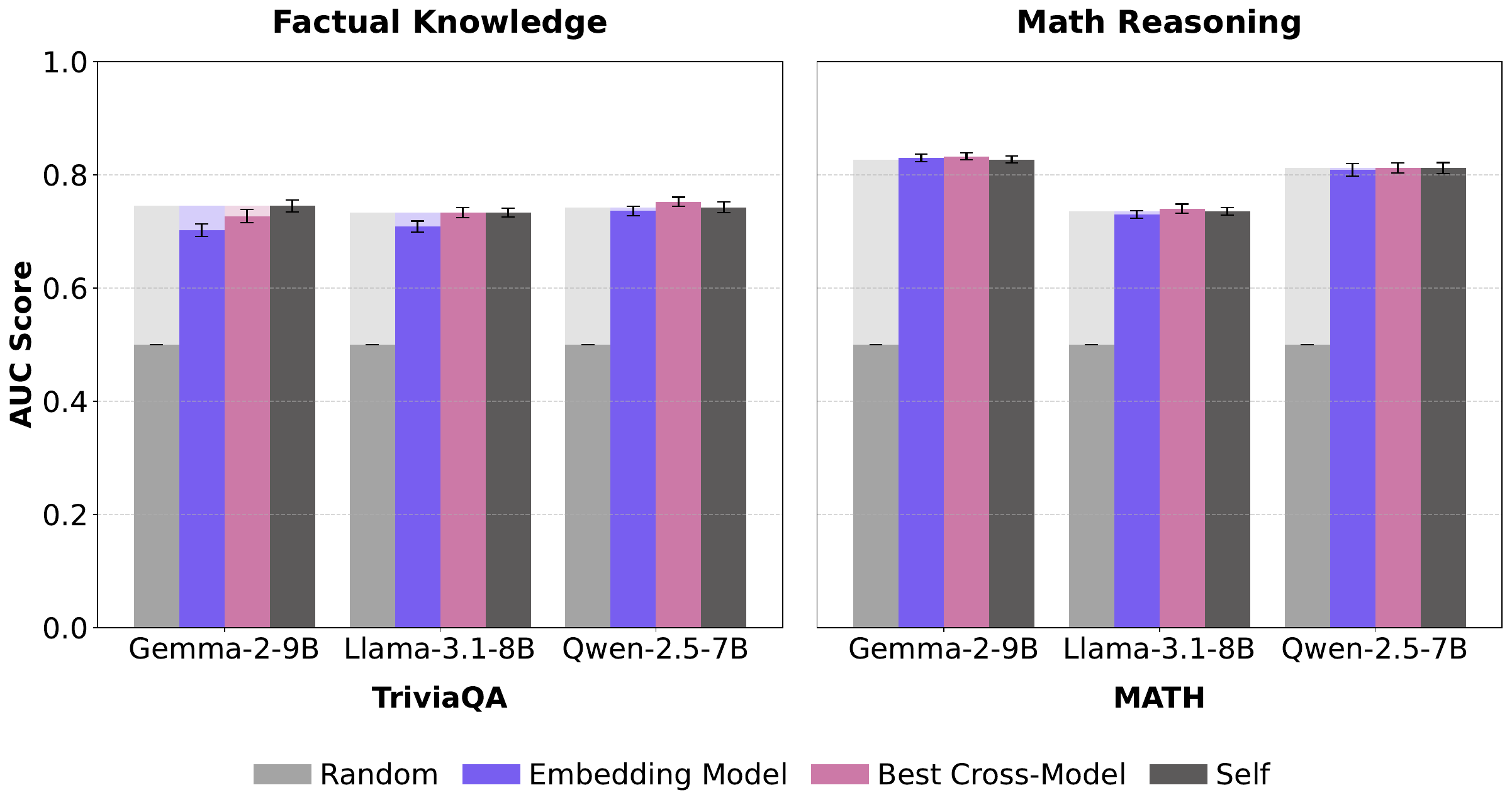}
    \caption{\textbf{Premium Gap.}
    Mean AUC for correctness prediction, averaged over layers, on two task types: factual knowledge (TriviaQA) and mathematical reasoning (MATH). Bars compare Random, Embedding, and Best Cross-Model baselines to the Self-Probe (\textit{Self}) across three target models. Semi-transparent overlays indicate the performance gain (or lack thereof) of \textit{Self} relative to each baseline. Error bars denote 95\% confidence intervals.}
\label{fig:premium_gap}
\end{figure*}

\section{Related Work}

\paragraph{LLM Introspection.} Recent work investigates whether models possess privileged access to their internal processes, often termed introspection. \citet{li2025training} find that models fine-tuned to explain their own internal computations—such as feature encoding and causal structure—outperform external explainers, suggesting a unique capacity for self-explanation. Similarly, \citet{binder2025looking} define introspection as knowledge derived from internal states, showing that models trained on their own behavior predict their hypothetical choices more accurately than third-party models. However, the reliability of this access is debated. \citet{li2025natural} argue that verbalized explanations often reflect the model's parametric knowledge rather than a faithful decoding of internal states, succeeding on benchmarks even without access to internals. Furthermore, \citet{binder2025looking} note that the observed self-prediction advantage is often limited to simple settings and does not consistently generalize to out-of-distribution tasks.

\paragraph{Using Probes to Predict Correctness.} LLMs have been shown to possess a degree of self-evaluation ability—accurately estimating answer correctness on multiple-choice tasks \citep{kadavath2022language} and identifying unanswerable questions \citep{Yin2023Do}—motivating direct investigation of their hidden states for correctness signals. Linear probes trained on the hidden states of reasoning models have been used to verify intermediate reasoning steps and even predict answer correctness prior to generation \citep{Zhang2025ReasoningMK}. Similarly, \citet{tamoyan2025factual} demonstrate that residual-stream features encode a ``factual self-awareness'' signal: simple linear projections can predict whether a model will recall a fact correctly. \citet{orgad2025llms} also report that internal representations carry rich truthfulness information, concentrated in specific tokens; notably, they show that a model may encode the correct answer internally even when its generated output is incorrect.

However, other studies argue that predictive probes often exploit artifacts from the question or answer rather than reflecting genuine introspection. \citet{Seo2025Quantifying} show that much of the reported probe accuracy arises from superficial question patterns. Extending this idea, \citet{xiao2025generalized} propose a `generalized correctness model' across multiple LLMs, finding that predictors trained on historical answer patterns perform comparably to model-specific probes, suggesting that LLMs have little privileged knowledge of their own correctness. This view is further supported by mechanistic analyses from \citet{Chi2025LargeLM}, who show that when LLMs hallucinate due to retrieving incorrect knowledge, their internal states are indistinguishable from those of correct answers, suggesting that LLMs do not explicitly encode correctness.

Our objective is to reconcile these two often conflicting lines of work. To this end, we employ a rigorous experimental framework designed to uncover whether LLMs genuinely possess privileged knowledge of their own correctness.

\section{Methodology}

\subsection{Problem Formulation}
\label{sec:pk-definition}

Let $M_{target}$ be the model whose correctness we wish to predict. For a given question $q$, $M_{target}$ generates an answer with a binary correctness label $y \in \{0, 1\}$.
We predict $y$ using a classifier (probe) $f$ trained on the internal hidden states $\mathbf{h}$ of a source model $M_{source}$ processing the same question $q$:
\begin{equation}
\hat{y} = f(\mathbf{h}(q; M_{source})).
\end{equation}

\paragraph{Defining Privileged Knowledge.}
To understand the information captured by the probe $f$, we posit that the hidden state $\mathbf{h}(q; M_{source})$ encodes two distinct latent components:
\begin{equation}
\mathbf{h} \approx \mathbf{z}_{public} \oplus \mathbf{z}_{private}
\end{equation}
Here, $\mathbf{z}_{public}$ captures inherent features of the input question $q$ (e.g., domain, entity types) which are externally accessible to any model processing the question. In contrast, $\mathbf{z}_{private}$ captures internal states specific to $M_{source}$ (e.g., memory retrieval success, reasoning confidence). We define $\mathbf{z}_{private}$ as \emph{privileged knowledge}, henceforth referring exclusively to internal states predictive of correctness.

\subsection{Probing Configurations}
To isolate privileged knowledge, we vary the source model $M_{source}$ to create distinct configurations. 
In all cases, the probes are trained and tested on the correctness labels $y$ of the target model ($M_{target}$).

\paragraph{1. Self-Probe.}
We set $M_{source} = M_{target}$. The probe is trained on the model's own representations to predict $y$.

\paragraph{2. External-Probe.}
We set $M_{source} \neq M_{target}$. The probe is trained on an external model's representations to predict $y$. 

We evaluate two types of external probes:
\begin{itemize}[itemsep=0pt, topsep=2pt]
    \item \textbf{Cross-Model:} $M_{source}$ is a peer LLM of comparable size (e.g., predicting Qwen's correctness using Llama's hidden states).
    \item \textbf{Embedding-Model:} $M_{source}$ is an embedding model of comparable size.
\end{itemize}

\subsection{Analysis Framework}

\paragraph{The Premium Gap.}
We refer to the advantage in correctness prediction performance of a self-probe over an external-probe as the \emph{premium gap}.
If privileged knowledge ($\mathbf{z}_{\text{private}}$) contains no correctness signal, then correctness prediction relies solely on public features ($\mathbf{z}_{\text{public}}$). In this case, external models with more informative representations of public features should outperform self-probes. 
Conversely, if a premium gap persists—where self-probes outperform all external probes—this provides evidence that the model possesses unique internal signals inaccessible to external observers.

\paragraph{Disagreement Subsets.}
On random samples, high inter-model agreement allows external probes to exploit peer correctness patterns as a proxy for the target model's behavior, masking any privileged knowledge signal.
To eliminate this confound, we evaluate performance on the disagreement subset, defined as the set of examples where $M_{target}$ and $M_{source}$ produce opposite correctness labels ($y_{\text{target}} \neq y_{\text{source}}$).
Crucially, we do \emph{not} retrain probes on this subset. Training exclusively on disagreement subsets would introduce a perfect negative correlation between self and external labels, allowing the probe to trivially exploit the external model's inverted correctness signal. Instead, we train probes on the full training dataset to learn the full correctness pattern of the source model, and filter predictions during inference to strictly evaluate on the disagreement test subset.

\subsection{Experimental Setup}
\label{sec:exp_setup}

\textbf{Models.}
We evaluate three instruction-tuned decoder LMs of comparable size: \texttt{Llama-3.1-8B}~\citep{grattafiori2024llama3herdmodels}, \texttt{Qwen2.5-7B}~\citep{qwen2025qwen25technicalreport}, and \texttt{Gemma-2-9B}~\citep{gemmateam2024gemma2improvingopen}, alongside the embedding model \texttt{Qwen3-Embedding-8B}~\citep{yang2025qwen3technicalreport}. The three decoder LMs serve as both target and source models, while the embedding model is used only as a source. To assess scalability, we additionally evaluate \texttt{Qwen-3-32B}~\citep{yang2025qwen3} as both a target model and an external probe candidate; results are reported in \Cref{app:qwen32b}.

\textbf{Datasets.} Our evaluation spans five datasets. Three focus on factual knowledge recall: Mintaka~\citep{sen2022mintaka}, TriviaQA~\citep{joshi2017triviaqa}, and HotpotQA~\citep{yang2018hotpotqa}. Two focus on mathematical reasoning: MATH~\citep{hendrycks2measuring} and GSM1K~\citep{zhang2024careful}. Note that while HotpotQA is often considered a multi-hop reasoning dataset, we use question-only evaluation without supporting documents, making it a test of parametric memory retrieval. See \Cref{app:dataset_sizes} for dataset sizes and \Cref{app:dataset_details} for evaluation protocols.

\textbf{Probing Method.} We extract hidden states of the question from the final token of every $5^{\text{th}}$ layer. Our primary analysis uses a \textbf{linear probe} (logistic regression with $L_2$ regularization). To ensure findings are not artifacts of linearity, we replicated all experiments using non-linear \textbf{MLP probes}, yielding qualitatively similar results (see \Cref{app:mlp_probe_results}). All probes are evaluated via Nested Stratified K-Fold Cross-Validation ($k=10$), reporting AUC on the aggregated out-of-fold probabilities.

\subsection{Evaluation Metrics}
We evaluate performance using the Area Under the ROC Curve (AUC). AUC is threshold-independent and robust to class imbalance, ensuring we measure genuine separation ability given the varying correctness rates across datasets.

\paragraph{Statistical Significance.}
We assess the significance of the premium gap using paired $t$-tests across validation folds. To control for family-wise error rates in multiple comparisons, we apply the Bonferroni-Holm correction \citep{holm1979simple} ($p < 0.05$).

\section{Results}

We present our empirical findings in three parts. First, we demonstrate that external-probes match self-probe performance in $2$ out of $3$ models in factual tasks and in all models in mathematical reasoning tasks (\Cref{sec:full_results}). Second, we identify high inter-model agreement on correctness labels as a critical confound: when models frequently agree on which questions they answer correctly or incorrectly, external probes can exploit the external model's own correctness patterns to predict the target model's behavior (\Cref{sec:agreement_hypothesis}). Third, by isolating performance on disagreement subsets, we reveal that a statistically significant yet modest premium gap emerges in factual tasks but remains absent in mathematical reasoning (\Cref{sec:disagreement_method}). We additionally evaluate on a larger model (\texttt{Qwen-3-32B}), verifying that the same overall trends hold despite its richer representations strengthening the external probe baseline (\Cref{app:qwen32b}).

\subsection{Full Test Sets Reveal No Premium Gap}\label{sec:full_results}

We first evaluate correctness prediction on the standard full test sets. As shown in \Cref{fig:premium_gap} (remaining datasets in \Cref{fig:premium_gap_rest}), self-probes successfully predict correctness across both factual knowledge and mathematical reasoning.
However, this performance is not unique to the model's internal states. In factual tasks, self-probes show only a small advantage over embedding model probes, and are comparable to cross-model probes in $2$ out of $3$ models. In mathematical reasoning, \emph{both} embedding model and cross-model probes match self-probe performance, yielding a non-existent premium gap.
This initial finding suggests that correctness prediction does not benefit from access to a model's unique internal states. The observed performance parity aligns with recent work by \citet{xiao2025generalized}, challenging the existence of privileged knowledge regarding a model's own correctness.

\begin{figure}[t]
    \centering
    \includegraphics[width=\linewidth]{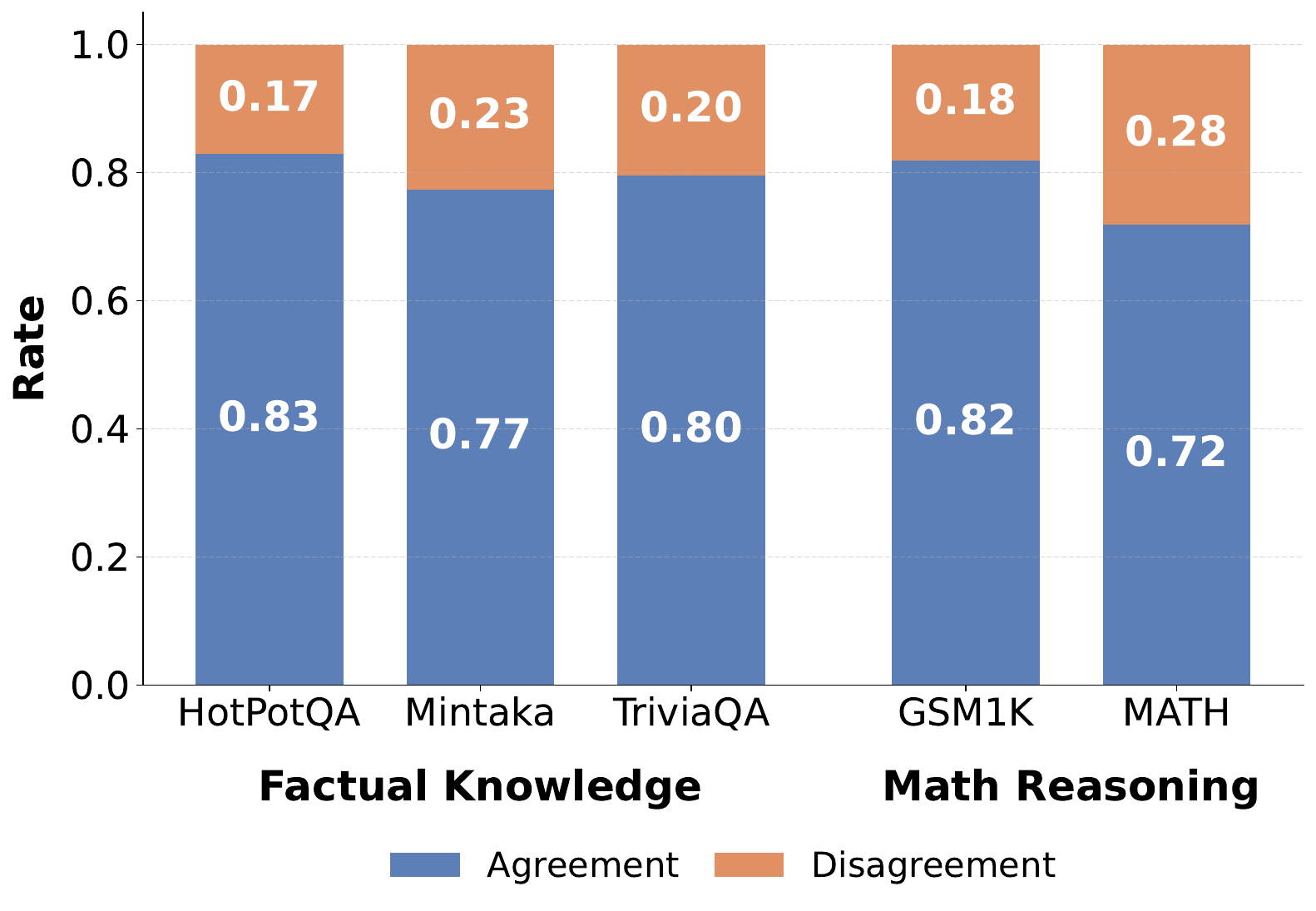}
    \caption{\textbf{Agreement vs. Disagreement Rates.}
    Stacked bar chart showing the proportion of questions on which models agree (blue) or disagree (orange) on correctness across datasets, averaged over all model pairs.}
    \label{fig:agreement_rate}
\end{figure}

\subsection{The Agreement Confound}
\label{sec:agreement_hypothesis}

We hypothesize that the absence of a premium gap in the cross-model scenario stems from a fundamental confound: \emph{inter-model agreement}. As shown in \Cref{fig:agreement_rate}, models agree on correctness for approximately $80\%$ of questions in factual tasks and $75\%$ in mathematical reasoning. This high agreement rate creates a critical problem for interpreting our results.

When models agree on the majority of examples, the external model's correctness becomes highly correlated with the target's correctness. This means that any signal in the external representation that predicts the external model's success—whether from input question features ($\mathbf{z}_{\text{public}}^{\text{ext}}$) or the external model's own privileged information about correctness ($\mathbf{z}_{\text{private}}^{\text{ext}}$)—will also predict the target's success on most cases. Consequently, an external probe can achieve high predictive performance compared to the self-probe without accessing the target model's privileged information.

This confound makes interpreting our full-set results particularly challenging. Concretely, in our experiments we consistently observe that Gemma representations dominate cross-model prediction: in linear probes, Gemma achieves the best performance as an external representation in 7 out of 9 factual cases (\Cref{fig:heatmap_results}), while in MLP probes, it is universally dominant across all 9 cases (\Cref{fig:heatmap_mlp_results}). However, this dominance is ambiguous and could reflect two very different underlying mechanisms:

\begin{enumerate}
    \item \textbf{No Privileged Knowledge:} Models truly lack internal correctness signals. Gemma simply encodes superior $\mathbf{z}_{\text{public}}$ features regarding question difficulty.
    \item \textbf{Masked Privileged Knowledge:} Privileged knowledge exists in both models, but Gemma's representation masks the target's signal. Because Gemma provides both robust public features \emph{and} a private signal that serves as a high-fidelity proxy (due to agreement), the superior summed contribution ($\mathbf{z}_{\text{public}}^{\text{Gemma}} \oplus \mathbf{z}_{\text{private}}^{\text{Gemma}}$) dominates the probe's learned weights, effectively obscuring the target's own internal signal.
\end{enumerate}

To distinguish between these scenarios, we evaluate on the \emph{Disagreement Subset}, where the external model's private signal cannot be leveraged to predict target correctness.

\subsection{Emergence of Domain-Specific Privileged Knowledge}
\label{sec:disagreement_method}

Evaluation on the disagreement subset reveals a sharp contrast between factual and mathematical domains across both linear and MLP probes (\Cref{fig:heatmap_results}, \Cref{fig:heatmap_mlp_results}).

\paragraph{Factual Knowledge: The Gap Emerges.}
In factual tasks, stripping away the agreement confound reveals genuine privileged knowledge. Self-probes consistently outperform external probes, exhibiting a statistically significant premium gap across all 9 experimental configurations using both linear and MLP probes (\Cref{fig:heatmap_results,fig:heatmap_mlp_results}, Left). This indicates that when the external proxy fails, external representations cannot fully account for the target model's correctness. The target model retains unique internal signals that remain inaccessible to observers. Detailed AUC scores for each factual dataset and self-cross model pairing are presented in \Cref{fig:disagreement_dumbbell_factual_lr,fig:disagreement_dumbbell_factual_mlp} (\Cref{app:detailed_disagreement_performance}).
While self-probes consistently outperform external-probes across all factual datasets and model pairs, AUC values on the disagreement subset are substantially lower than on the full test set. This is expected, as inter-model disagreement indicates boundary regions where models exhibit higher uncertainty \citep{lakshminarayanan2017simple}, resulting in less stable correctness patterns that are harder to predict.
We discuss and argue against a potential alternative explanation 
based on distributional shifts in \Cref{app:distributional_shift}.

\paragraph{Mathematical Reasoning: No Evidence for Privileged Knowledge.}
In sharp contrast, mathematical reasoning tasks show no premium gap. Even on the disagreement subset, external model probes closely match or outperform self-probes across all targets (\Cref{fig:heatmap_results,fig:heatmap_mlp_results}, Right).
Detailed AUC scores for each mathematical reasoning dataset and model pairing are presented in \Cref{fig:disagreement_dumbbell_math_lr,fig:disagreement_dumbbell_math_mlp} (\Cref{app:detailed_disagreement_performance}).

\section{Where Does Privileged Knowledge Emerge?}
\label{sec:analysis}

Our findings in \Cref{sec:disagreement_method} establish that privileged knowledge emerges in factual tasks but remains absent in mathematical reasoning, based on layer-averaged premium gaps. A natural follow-up question is \textit{where} in the network this signal originates: is the premium gap uniformly distributed across layers, or does it emerge at specific depths?

To investigate this, we compute the premium gap (self-probe AUC $-$ best external-probe AUC) at each individually probed layer on the disagreement subset, rather than averaging across layers as in \Cref{sec:disagreement_method}. We probe every $5^{\text{th}}$ layer plus the final layer, yielding 6--7 measurement points per model, and plot the premium gap against normalized layer depth (0 = first probed layer, 1 = last). 
Per-model bar figures showing absolute self and external AUC values at each layer are provided in \Cref{app:layer_analysis_figures}.

\subsection{Factual Tasks: Progressive Emergence}

\begin{figure*}[t]
    \centering
    \begin{subfigure}[t]{\linewidth}
        \centering
        \includegraphics[width=\linewidth]{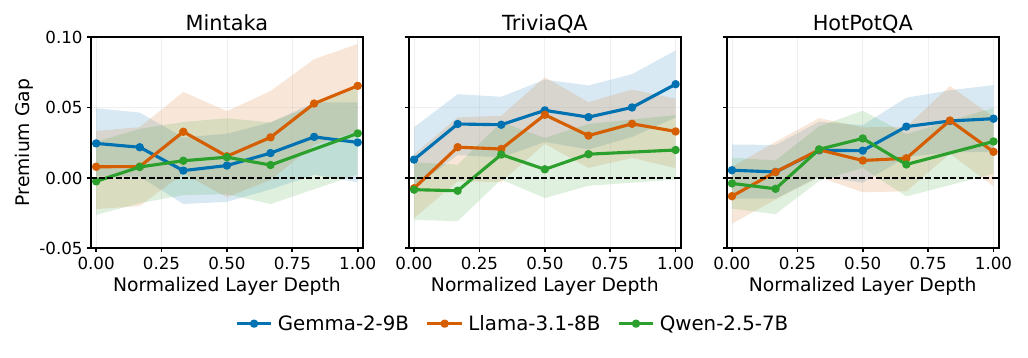}
        \caption{Factual Knowledge.}
        \label{fig:layer_premium_gap_factual}
    \end{subfigure}
    \vspace{0.5em}
    \begin{subfigure}[t]{\linewidth}
        \centering
        \includegraphics[width=\linewidth]{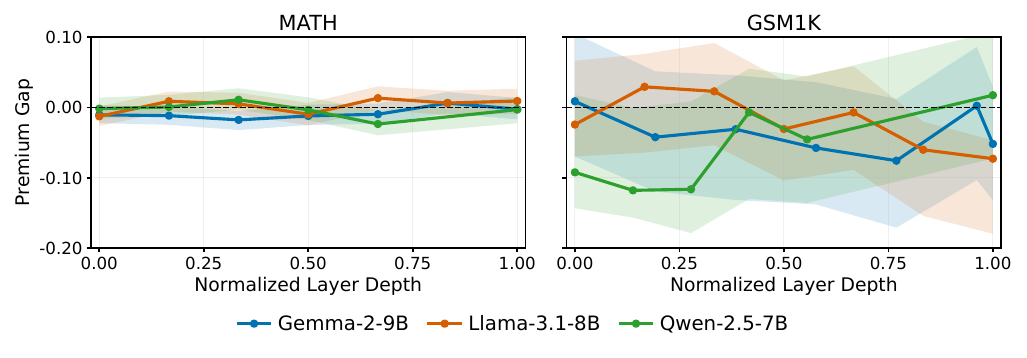}
        \caption{Mathematical Reasoning.}
        \label{fig:layer_premium_gap_math}
    \end{subfigure}
    \caption{\textbf{Per-Layer Premium Gap.}
    Premium gap (self-probe AUC $-$ best external-probe AUC) on the disagreement subset as a function of normalized layer depth. Shaded regions denote 95\% confidence intervals.
    \textbf{(a)}~For factual tasks, the gap is near zero in early layers and grows progressively toward deeper layers across all models, indicating that privileged knowledge emerges in early-to-mid representations.
    \textbf{(b)}~For mathematical reasoning, MATH fluctuates near zero at all depths and GSM1K is predominantly negative (external probes outperform self-probes). No layer exhibits a consistent self-probe advantage across the full network depth.}
\end{figure*}

\Cref{fig:layer_premium_gap_factual} presents the per-layer premium gap for factual datasets. Across all three models and datasets, a consistent pattern emerges: the premium gap is near zero or slightly negative in early layers and grows progressively toward deeper layers.
Early layers, which primarily encode surface-level and syntactic features \citep{tenney2019bert,jawahar2019does}, show no self-probe advantage,
 consistent with these representations encoding primarily public information ($\mathbf{z}_{\text{public}}$). The premium gap becomes reliably positive from approximately layer~10--15 onward (normalized depth $\sim$0.25--0.40), suggesting that privileged knowledge is encoded in deeper representations where models consolidate factual knowledge~\citep{orgad2025llms}.
This pattern is consistent with the view that the privileged signal reflects idiosyncratic memory retrieval states that build up through the forward pass. In particular, \citet{Chi2025LargeLM} show that knowledge recall in LLMs is dominated by a mid-layer information-flow signal from subject to answer tokens, aligning with our finding that the privileged advantage first appears in early-to-mid layers and strengthens toward later layers.

\subsection{Mathematical Reasoning: No Consistent Advantage}

The per-layer analysis for mathematical reasoning (\Cref{fig:layer_premium_gap_math}) confirms that the absence of privileged knowledge in math is not an artifact of layer averaging. For MATH, the premium gap fluctuates near zero across all layers and models, with no systematic trend. For GSM1K, the gap is predominantly \textit{negative}---external probes outperform self-probes at most depths, inverting the factual pattern.

This layer-level analysis strengthens the domain-specificity conclusion from \Cref{sec:disagreement_method}. Mathematical correctness signals appear to be publicly accessible at every depth of the network, suggesting that reasoning difficulty is governed by problem structure rather than model-specific knowledge. 

\subsection{What Drives the Correctness Signal?}
The layer-level analysis above localizes where privileged knowledge emerges, but does not address what linguistic features drive correctness prediction more broadly. In \Cref{app:lexical_only}, we strip questions of their syntax to isolate named entities and nouns, finding that concept-level familiarity accounts for a substantial portion of the predictive performance in factual tasks and MATH, whereas GSM1K correctness relies on structural problem features that lexical stripping destroys.

\section{Discussion}

We investigate whether LLMs possess privileged knowledge about the correctness of their forthcoming answer by comparing probes trained on self-representations versus external model representations. Our key methodological contribution is identifying inter-model agreement as a critical confound: when models share correctness patterns, external probes exploit peer correctness as a proxy, masking genuine privileged signals. Evaluating on disagreement subsets reveals that privileged knowledge is domain-specific --- it emerges consistently in factual tasks, while mathematical reasoning correctness remains externally observable.

These findings reconcile prior conflicting results: privileged knowledge does exist, but is domain-specific and was previously masked by inter-model agreement. Beyond correctness prediction, our disagreement-based methodology can be extended to study privileged knowledge in hybrid domains (coding, commonsense reasoning) and other forms of model introspection. Practically, our results suggest that model-specific activations carry signals that black-box tools miss, with potential applications in model monitoring.

Our probe-based analysis is correlational in nature, and the causal mechanisms underlying the privileged knowledge signal remain an open question. A natural avenue for future work is activation steering: if the factual correctness signal is genuinely tied to subject-specific retrieval, intervening on the identified correctness direction in the residual stream should predictably modulate output correctness.

\section{Limitations}
\label{sec:limitations}
Our analysis has several limitations: (1) Although we additionally evaluate \texttt{Qwen-3-32B} (\Cref{app:qwen32b}), our main analysis is limited to models with $7$B--$9$B parameters; larger models may display different patterns of privileged knowledge. (2) Our scope is limited to factual knowledge and mathematical reasoning, while hybrid domains such as coding and commonsense reasoning remain outside the scope of this study. (3) We rely on linear and MLP probes which, although standard in prior work, may have limited capacity to fully extract privileged signals. (4) Our study reveals systematic patterns linking representational structure to correctness; complementary intervention experiments could further establish the causal mechanisms underlying factual privileged knowledge.

\section*{Acknowledgements}
This research is funded by the European Union (ERC, Control-LM, 101165402) and supported by the Israel Science Foundation (grant No. 2942/25). Views and opinions expressed are those of the authors only and do not necessarily reflect those of the European Union or the European Research Council Executive Agency. Neither the European Union nor the granting authority can be held responsible for them.

\bibliography{custom}

\clearpage

\appendix
\label{sec:appendix}

\section{Additional Experimental Results}
\label{app:additional_results}

\subsection{Larger Model: Qwen-3-32B}
\label{app:qwen32b}

To assess scalability, we evaluate \texttt{Qwen-3-32B} using the same 
pipeline as the main experiments, incorporating it both as a 
target model and as an external probe candidate for the three 
main models. As a significantly larger model, \texttt{Qwen-3-32B} likely 
produces richer public feature representations, which would be 
expected to benefit external probes—making it a stronger test 
of our claims, while also strengthening the best cross-model 
baseline for the three main models; results therefore differ 
slightly from \Cref{fig:heatmap_results}.

Layer-averaged and best-layer results are presented in 
\Cref{tab:qwen32b} respectively, and are 
consistent with our main findings. On the full test set, no 
model exhibits a reliable premium gap, consistent with our 
inter-model agreement analysis (\Cref{sec:agreement_hypothesis}). 
On the disagreement subset, self-probe advantages emerge in 
factual tasks across all models, most consistently on TriviaQA 
and HotPotQA, while Mintaka shows weaker and less consistent 
gains. \texttt{Qwen-3-32B} follows the same pattern, with statistically 
significant advantages on TriviaQA and HotPotQA in both 
evaluation settings.

\begin{table*}[t]
\centering
\small
\setlength{\tabcolsep}{4pt}

\begin{subtable}{\textwidth}
\centering
\resizebox{\textwidth}{!}{%
\begin{tabular}{ll|ccc|cc}
\toprule
& \textbf{Target Model} 
& \textbf{Mintaka} & \textbf{TriviaQA} & \textbf{HotPotQA} 
& \textbf{MATH} & \textbf{GSM1K} \\
\midrule
\multirow{4}{*}{\textbf{Full}}
& Gemma-2-9B   
    & $\mathbf{+0.017}^\star$ \scriptsize{(6.5\%)}  
    & $\mathbf{+0.018}^\star$ \scriptsize{(6.7\%)}  
    & $\mathbf{+0.021}^\star$ \scriptsize{(8.9\%)}  
    & $-0.006$ \scriptsize{($-3.4\%$)}              
    & $-0.025$ \scriptsize{($-8.3\%$)}              \\
& Llama-3.1-8B 
    & $\mathbf{+0.002}$ \scriptsize{(0.8\%)}        
    & $\mathbf{+0.000}$ \scriptsize{(0.1\%)}        
    & $-0.012$ \scriptsize{($-5.6\%$)}              
    & $-0.005$ \scriptsize{($-1.9\%$)}              
    & $-0.014$ \scriptsize{($-4.8\%$)}              \\
& Qwen-2.5-7B  
    & $-0.005$ \scriptsize{($-2.0\%$)}              
    & $-0.010$ \scriptsize{($-4.0\%$)}              
    & $-0.007$ \scriptsize{($-3.6\%$)}              
    & $-0.000$ \scriptsize{($-0.1\%$)}              
    & $-0.011$ \scriptsize{($-3.8\%$)}              \\
& Qwen-3-32B   
    & $-0.014$ \scriptsize{($-5.6\%$)}              
    & $-0.010$ \scriptsize{($-3.6\%$)}              
    & $-0.006$ \scriptsize{($-3.0\%$)}              
    & $-0.013$ \scriptsize{($-5.2\%$)}              
    & $-0.020$ \scriptsize{($-4.5\%$)}              \\
\midrule
\multirow{4}{*}{\textbf{Disagree}}
& Gemma-2-9B   
    & $\mathbf{+0.023}$ \scriptsize{(5.1\%)}        
    & $\mathbf{+0.034}^\star$ \scriptsize{(6.8\%)}  
    & $\mathbf{+0.025}^\star$ \scriptsize{(4.8\%)}  
    & $-0.010$ \scriptsize{($-3.0\%$)}              
    & $-0.133$ \scriptsize{($-36.6\%$)}             \\
& Llama-3.1-8B 
    & $\mathbf{+0.032}^\star$ \scriptsize{(6.4\%)}  
    & $\mathbf{+0.018}^\star$ \scriptsize{(3.5\%)}  
    & $\mathbf{+0.020}^\star$ \scriptsize{(4.2\%)}  
    & $-0.011$ \scriptsize{($-1.8\%$)}              
    & $-0.090$ \scriptsize{($-24.2\%$)}             \\
& Qwen-2.5-7B  
    & $\mathbf{+0.016}$ \scriptsize{(3.2\%)}        
    & $\mathbf{+0.015}^\star$ \scriptsize{(3.4\%)}  
    & $\mathbf{+0.015}^\star$ \scriptsize{(3.4\%)}  
    & $-0.004$ \scriptsize{($-0.9\%$)}              
    & $-0.173$ \scriptsize{($-51.0\%$)}             \\
& Qwen-3-32B   
    & $-0.000$ \scriptsize{($-0.1\%$)}              
    & $\mathbf{+0.022}^\star$ \scriptsize{(4.1\%)}  
    & $\mathbf{+0.015}^\star$ \scriptsize{(3.2\%)}  
    & $-0.009$ \scriptsize{($-2.0\%$)}              
    & $-0.024$ \scriptsize{($-4.4\%$)}              \\
\bottomrule
\end{tabular}%
}
\caption{Averaged over all layers}
\end{subtable}

\bigskip

\begin{subtable}{\textwidth}
\centering
\resizebox{\textwidth}{!}{%
\begin{tabular}{ll|ccc|cc}
\toprule
& \textbf{Target Model} 
& \textbf{Mintaka} & \textbf{TriviaQA} & \textbf{HotPotQA} 
& \textbf{MATH} & \textbf{GSM1K} \\
\midrule
\multirow{4}{*}{\textbf{Full}}
& Gemma-2-9B   
    & $\mathbf{+0.011}$ \scriptsize{(4.5\%)}        
    & $\mathbf{+0.031}^\star$ \scriptsize{(12.5\%)} 
    & $\mathbf{+0.024}^\star$ \scriptsize{(11.1\%)} 
    & $-0.004$ \scriptsize{($-2.9\%$)}              
    & $-0.028$ \scriptsize{($-10.3\%$)}             \\
& Llama-3.1-8B 
    & $-0.011$ \scriptsize{($-4.5\%$)}              
    & $-0.007$ \scriptsize{($-2.9\%$)}              
    & $-0.012$ \scriptsize{($-5.9\%$)}              
    & $-0.010$ \scriptsize{($-4.3\%$)}              
    & $-0.009$ \scriptsize{($-3.3\%$)}              \\
& Qwen-2.5-7B  
    & $\mathbf{+0.009}$ \scriptsize{(3.7\%)}        
    & $-0.005$ \scriptsize{($-2.1\%$)}              
    & $\mathbf{+0.000}$ \scriptsize{(0.2\%)}        
    & $\mathbf{+0.004}$ \scriptsize{(2.3\%)}        
    & $-0.029$ \scriptsize{($-11.8\%$)}             \\
& Qwen-3-32B   
    & $-0.005$ \scriptsize{($-2.0\%$)}              
    & $\mathbf{+0.001}$ \scriptsize{(0.5\%)}        
    & $-0.003$ \scriptsize{($-1.6\%$)}              
    & $-0.016$ \scriptsize{($-6.6\%$)}              
    & $-0.018$ \scriptsize{($-4.5\%$)}              \\
\midrule
\multirow{4}{*}{\textbf{Disagree}}
& Gemma-2-9B   
    & $\mathbf{+0.056}^\star$ \scriptsize{(11.7\%)} 
    & $\mathbf{+0.081}^\star$ \scriptsize{(15.5\%)} 
    & $\mathbf{+0.035}^\star$ \scriptsize{(6.9\%)}  
    & $\mathbf{+0.005}$ \scriptsize{(1.8\%)}        
    & $-0.145$ \scriptsize{($-45.2\%$)}             \\
& Llama-3.1-8B 
    & $\mathbf{+0.065}^\star$ \scriptsize{(12.6\%)} 
    & $\mathbf{+0.030}^\star$ \scriptsize{(6.0\%)}  
    & $\mathbf{+0.042}^\star$ \scriptsize{(8.3\%)}  
    & $-0.007$ \scriptsize{($-1.3\%$)}              
    & $-0.090$ \scriptsize{($-26.8\%$)}             \\
& Qwen-2.5-7B  
    & $\mathbf{+0.033}$ \scriptsize{(7.0\%)}        
    & $\mathbf{+0.039}^\star$ \scriptsize{(8.6\%)}  
    & $\mathbf{+0.031}^\star$ \scriptsize{(6.9\%)}  
    & $\mathbf{+0.011}$ \scriptsize{(3.4\%)}        
    & $-0.096$ \scriptsize{($-24.4\%$)}             \\
& Qwen-3-32B   
    & $\mathbf{+0.005}$ \scriptsize{(1.2\%)}        
    & $\mathbf{+0.012}$ \scriptsize{(3.0\%)}        
    & $\mathbf{+0.045}^\star$ \scriptsize{(9.1\%)}  
    & $\mathbf{+0.001}$ \scriptsize{(0.1\%)}        
    & $\mathbf{+0.015}$ \scriptsize{(2.7\%)}        \\
\bottomrule
\end{tabular}%
}
\caption{Averaged over all layers}
\end{subtable}

\caption{
Correctness prediction results with Qwen-3-32B included.
Each cell reports $\Delta$AUC (Self $-$ Best Cross) with gap closed (\%) in parentheses.
\textbf{Bold} indicates self-probe advantage (positive $\Delta$AUC).
$^\star$: $p < 0.05$, Holm--Bonferroni corrected.
Results for Gemma-2-9B, Llama-3.1-8B, and Qwen-2.5-7B differ slightly from \Cref{fig:heatmap_results}
as Qwen-3-32B representations are included as an additional external probe candidate,
strengthening the best cross-model baseline.
}
\label{tab:qwen32b}

\end{table*}

\subsection{MLP Probe Results}
\label{app:mlp_probe_results}

To ensure that the vanishing premium gap is not an artifact of the limited expressivity of linear classifiers, we replicated our primary analysis using non-linear Multi-Layer Perceptron (MLP) probes (implementation details in Appendix~\ref{app:implementation_details}).

\begin{figure*}[t]
    \centering
    \includegraphics[width=\linewidth]{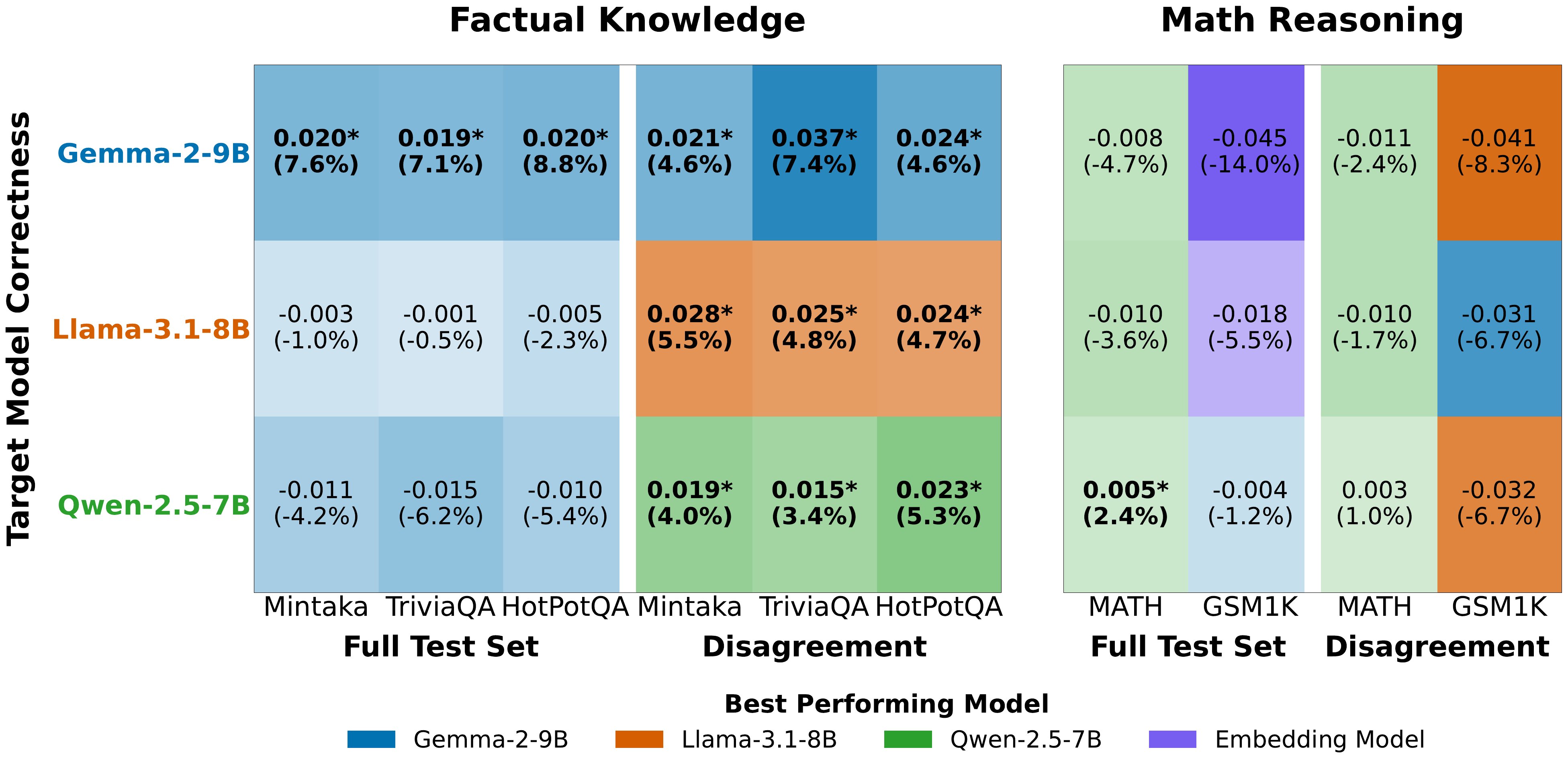}
    \caption{\textbf{Target Model Correctness Prediction (MLP Probes).}
    Heatmap of correctness prediction differences across target models, datasets, and test subsets. Each cell reports the AUC difference ($\Delta \mathrm{AUC} = \mathrm{Self} - \mathrm{Best\ External}$), with the percentage of the gap closed shown in parentheses, computed as $\frac{\mathrm{Self}-\mathrm{Best\ External}}{1-\mathrm{Best\ External}}\times 100$. The y-axis lists target models, and cell colors indicate the best-performing external source model for each setting. Asterisks (*) denote statistically significant differences (paired $t$-test, $p<0.05$, Bonferroni--Holm correction).}
    \label{fig:heatmap_mlp_results}
\end{figure*}

\begin{figure*}[t]
    \centering
    \begin{subfigure}[t]{\linewidth}
        \centering
        \includegraphics[width=0.9\linewidth]{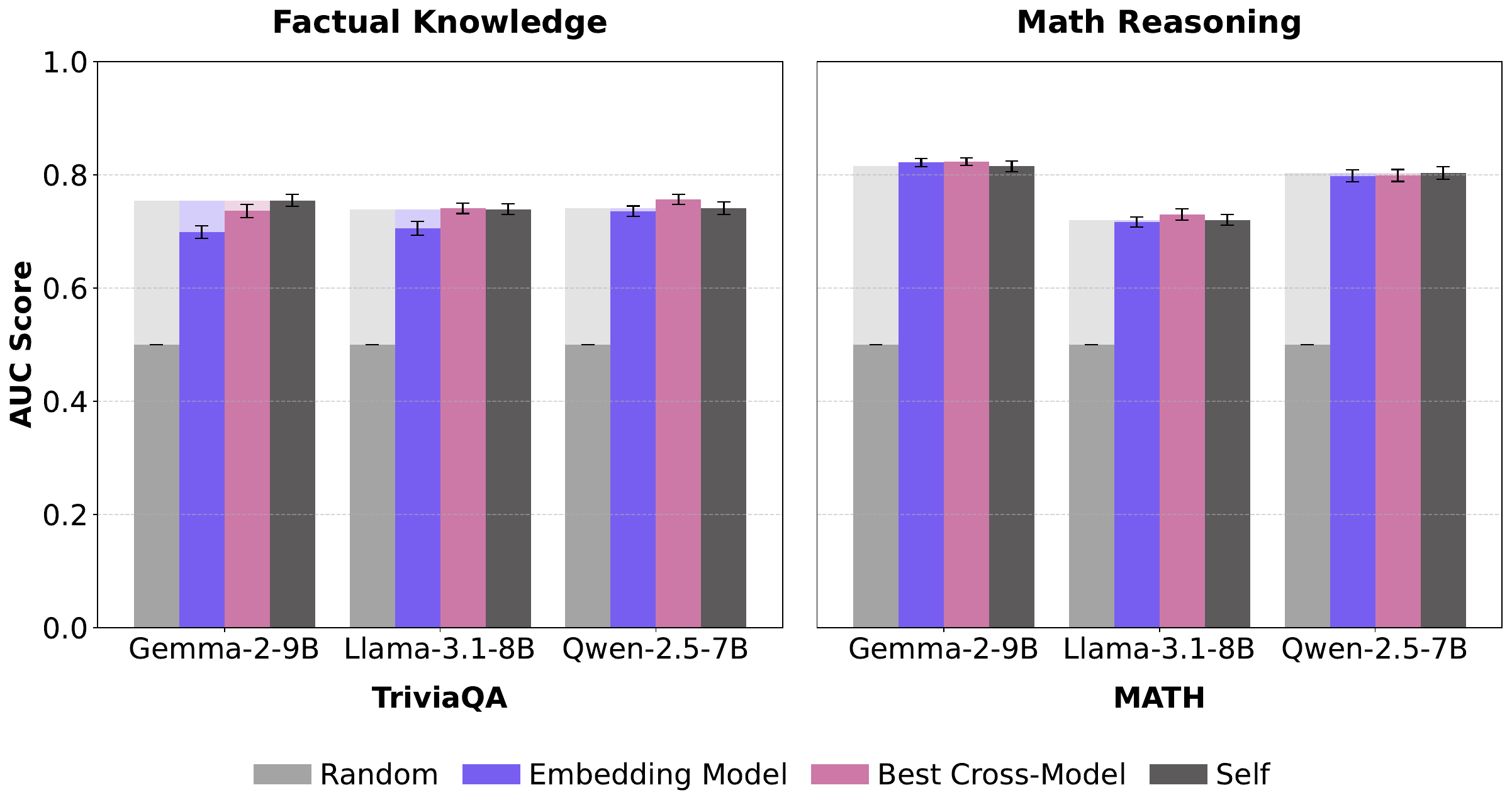}
        \caption{TriviaQA (Factual Knowledge) and MATH (Mathematical Reasoning).}
        \label{fig:premium_gap_mlp_main}
    \end{subfigure}
    \vspace{0.5em}
    \begin{subfigure}[t]{\linewidth}
        \centering
        \includegraphics[width=0.9\linewidth]{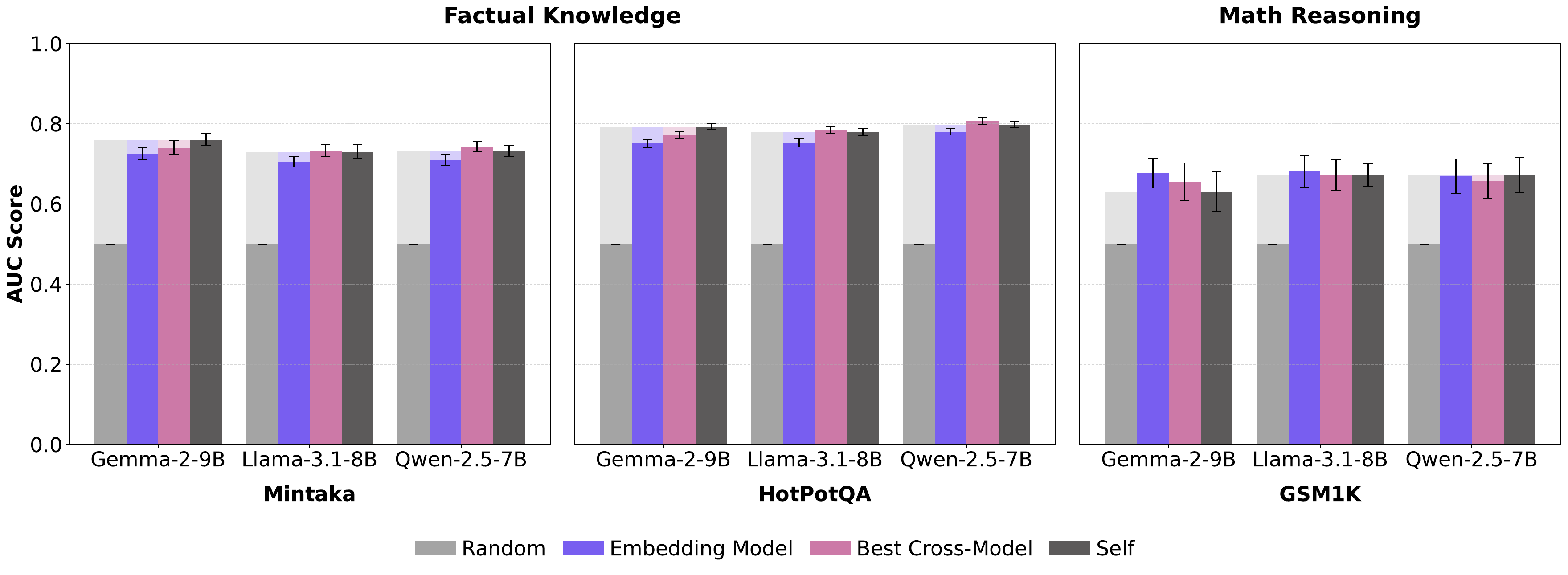}
        \caption{Remaining datasets: Mintaka, HotPotQA (Factual) and GSM1K (Math).}
        \label{fig:premium_gap_rest_mlp}
    \end{subfigure}
    \caption{\textbf{Premium Gap (MLP Probes).}
    Mean AUC for correctness prediction, averaged over layers. Bars compare Random, Embedding, and Best Cross-Model baselines to the Self-Probe (\textit{Self}) across three target models. Semi-transparent overlays indicate the performance gain (or lack thereof) of \textit{Self} relative to each baseline. Error bars denote 95\% confidence intervals.
    \textbf{(a)}~Core comparison on TriviaQA and MATH, mirroring the logistic-regression analysis in \Cref{fig:premium_gap}.
    \textbf{(b)}~Extended results on the remaining datasets.}
    \label{fig:premium_gap_all_mlp}
\end{figure*}

The results, visualized in \Cref{fig:premium_gap_all_mlp,fig:heatmap_mlp_results}, align closely with the linear probe findings, demonstrating that our conclusions are robust to probe architecture:

\paragraph{Full Test Set.} 
On the full test sets, the premium gap diminishes or vanishes, particularly in mathematical reasoning. Similar to the linear setting, we observe strong external dominance. Notably, Gemma representations are even more dominant in the non-linear setting, achieving the best cross-model performance in all 9 experimental configurations (~\Cref{fig:heatmap_mlp_results}). This reinforces the hypothesis that external representations often capture public correctness features more effectively than the target's own features.

\paragraph{Disagreement Subset.}
Crucially, the emergence of privileged knowledge in factual tasks is fully consistent across both probe types. Both linear and MLP probes detect a significant premium gap in all 9 factual configurations, confirming that the target model retains unique internal signals inaccessible to external observers. Conversely, in mathematical reasoning (GSM1K, MATH), the premium gap remains absent under both probe types, further validating that mathematical correctness signals are publicly accessible even to non-linear observers.

\section{Dataset and Disagreement Statistics}
\label{app:dataset_sizes}

We utilize five datasets with varying total sample sizes (Total). A critical component of our methodology involves analyzing the \textit{disagreement subset}—instances where the source and target models predict different correctness labels ($y_{\text{ext}} \neq y_{\text{target}}$). 

Since the size of this subset varies depending on the specific pair of models being compared, we report the exact counts ($N_{\text{disagreement}}$) for each unique model pair in \Cref{tab:dataset_sizes}. Across all configurations, the disagreement subsets retain sufficient scale ($\approx 20\%$ of the original data).

\begin{table}[h]
    \centering
    \small
    \resizebox{\columnwidth}{!}{%
    \begin{tabular}{l c c c c}
        \toprule
        & & \multicolumn{3}{c}{\textbf{Subset Size ($N_{\text{disagreement}}$)}} \\
        \cmidrule(lr){3-5}
        \textbf{Dataset} & \textbf{Total} & \textbf{G $\leftrightarrow$ L} & \textbf{G $\leftrightarrow$ Q} & \textbf{L $\leftrightarrow$ Q} \\
        \midrule
        \multicolumn{5}{l}{\textit{Mathematical Reasoning}} \\
        GSM1K & 1k & 186 & 142 & 216 \\
        MATH & 10k & 2,932 & 2,967 & 2,519 \\
        \midrule
        \multicolumn{5}{l}{\textit{Factual Knowledge}} \\
        HotpotQA & 10k & 1,592 & 1,730 & 1,802 \\
        Mintaka & 4k & 805 & 973 & 946 \\
        TriviaQA & 10k & 1,588 & 2,238 & 2,320 \\
        \bottomrule
    \end{tabular}%
    }
    \caption{Dataset statistics and disagreement subset sizes. Total denotes the full test set size. The subsequent columns show the size of the disagreement subset for each unique model pair. (G=Gemma-2-9B, L=Llama-3.1-8B, Q=Qwen2.5-7B).}
    \label{tab:dataset_sizes}
\end{table}

\begin{figure*}[t]
    \centering
    \includegraphics[width=\linewidth]{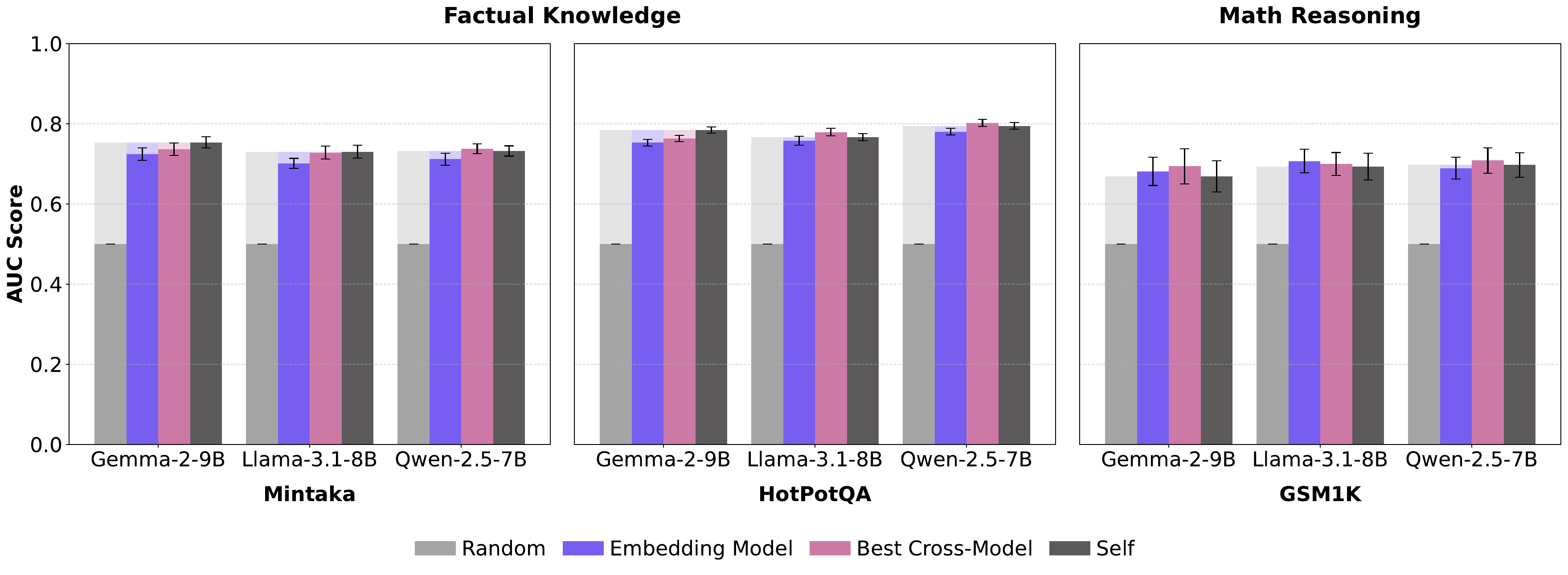}
    \caption{\textbf{Premium Gap (Remaining Datasets).}
    Mean AUC for correctness prediction, averaged over layers, on two task types: factual knowledge (Mintaka, HotPotQA) and mathematical reasoning (GSM1K). Bars compare Random, Embedding, and Best Cross-Model baselines to the Self-Probe (\textit{Self}) across three target models. Semi-transparent overlays indicate the performance gain (or lack thereof) of \textit{Self} relative to each baseline. Error bars denote 95\% confidence intervals.}
    \label{fig:premium_gap_rest}
\end{figure*}

\section{Detailed Disagreement Performance}
\label{app:detailed_disagreement_performance}

This section provides the detailed performance breakdowns on the disagreement subsets across all datasets, comparing self-probes against external probes for both linear and MLP configurations. 

\begin{figure*}[t]
    \centering
    \includegraphics[width=\linewidth]{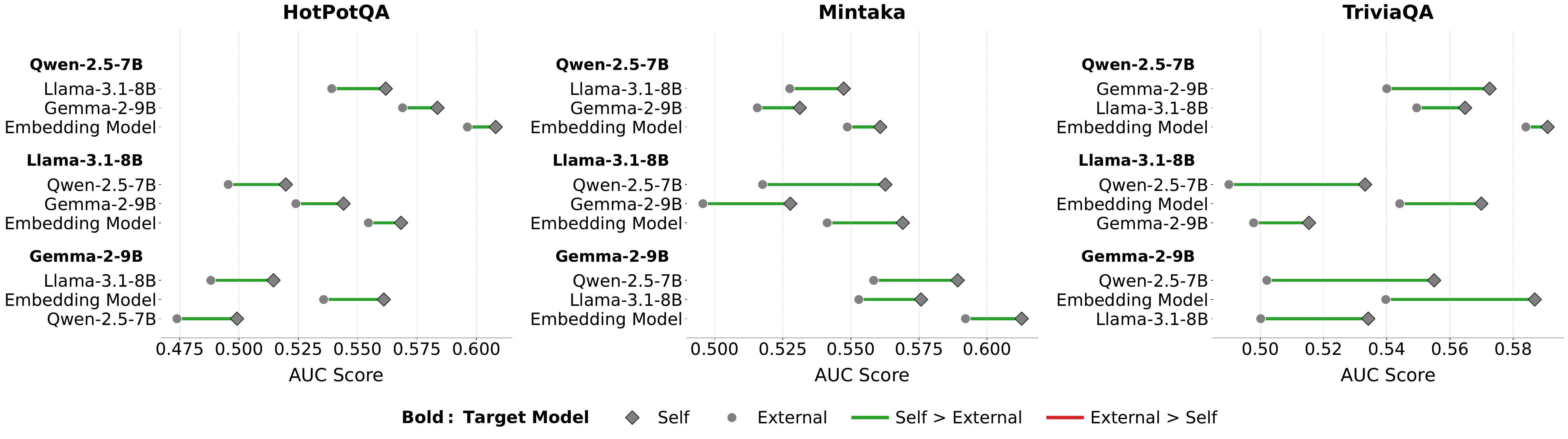}
    \caption{\textbf{Disagreement Gap: Factual Knowledge Breakdown (Linear Probes).}
    Detailed performance on the disagreement subset across Mintaka, TriviaQA, and HotPotQA.
    Self-probes consistently outperform external probes across all factual datasets, reinforcing the existence of privileged knowledge in factual tasks.}
    \label{fig:disagreement_dumbbell_factual_lr}
\end{figure*}

\begin{figure*}[t]
    \centering
    \includegraphics[width=\linewidth]{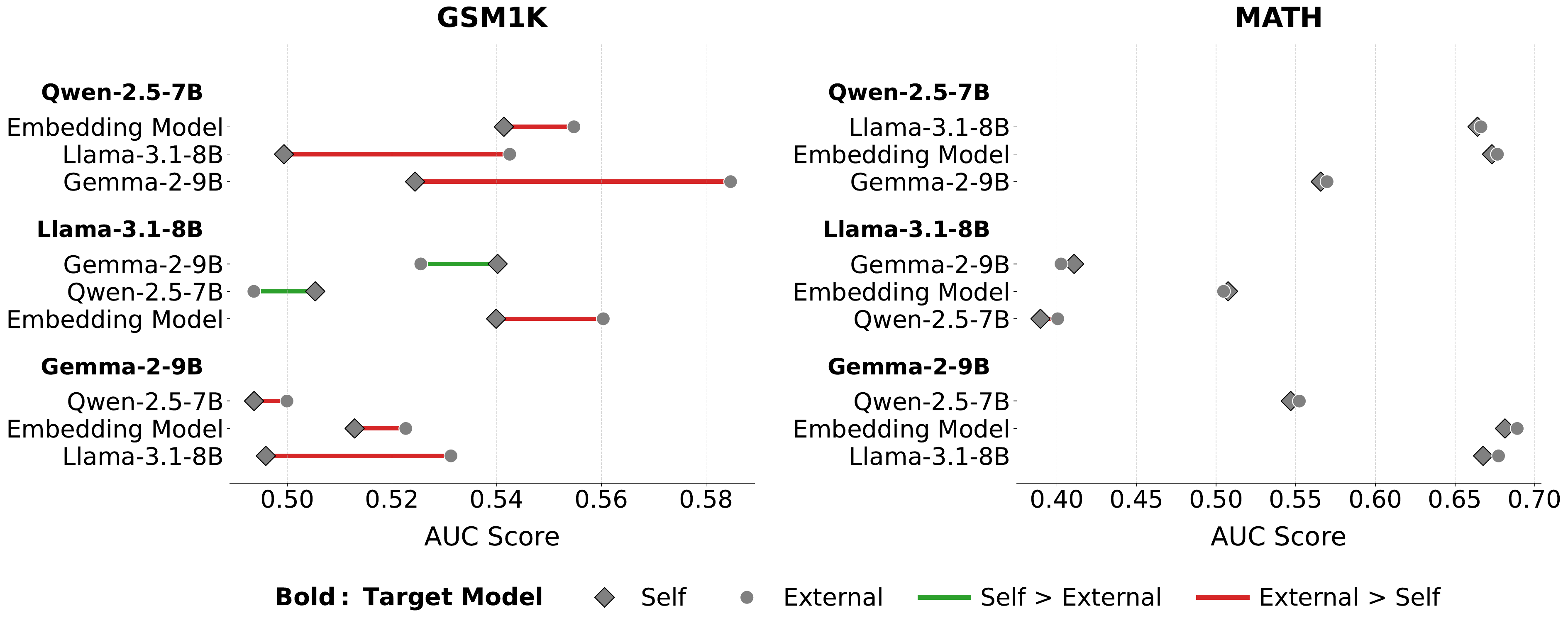}
    \caption{\textbf{Disagreement Gap: Mathematical Reasoning Breakdown (Linear Probes).}
    Detailed performance on the disagreement subset across GSM1K and MATH.
    Unlike factual tasks, mathematical correctness shows no consistent premium gap, indicating that reasoning difficulty is a public feature accessible to external models.}
    \label{fig:disagreement_dumbbell_math_lr}
\end{figure*}

\begin{figure*}[t]
    \centering
    \includegraphics[width=\linewidth]{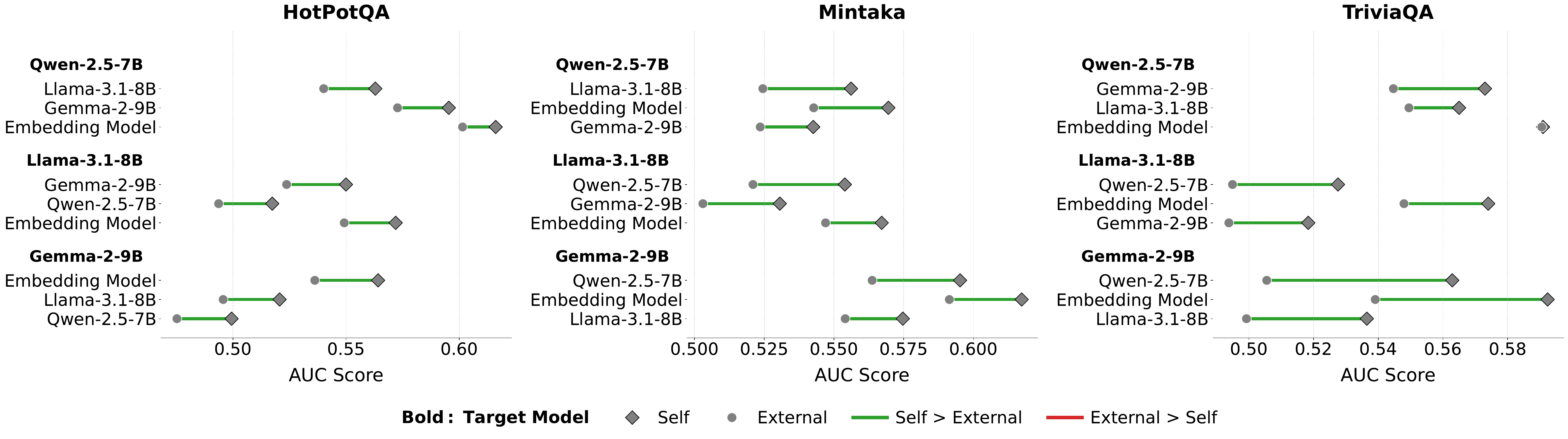}
    \caption{\textbf{Disagreement Gap: Factual Knowledge Breakdown (MLP Probes).}
    Detailed disagreement subset performance using MLP probes.
    The premium gap is even more pronounced with non-linear probes, with Self-representations outperforming Best External probes in 9 out of 9 configurations.}
    \label{fig:disagreement_dumbbell_factual_mlp}
\end{figure*}

\begin{figure*}[t]
    \centering
    \includegraphics[width=\linewidth]{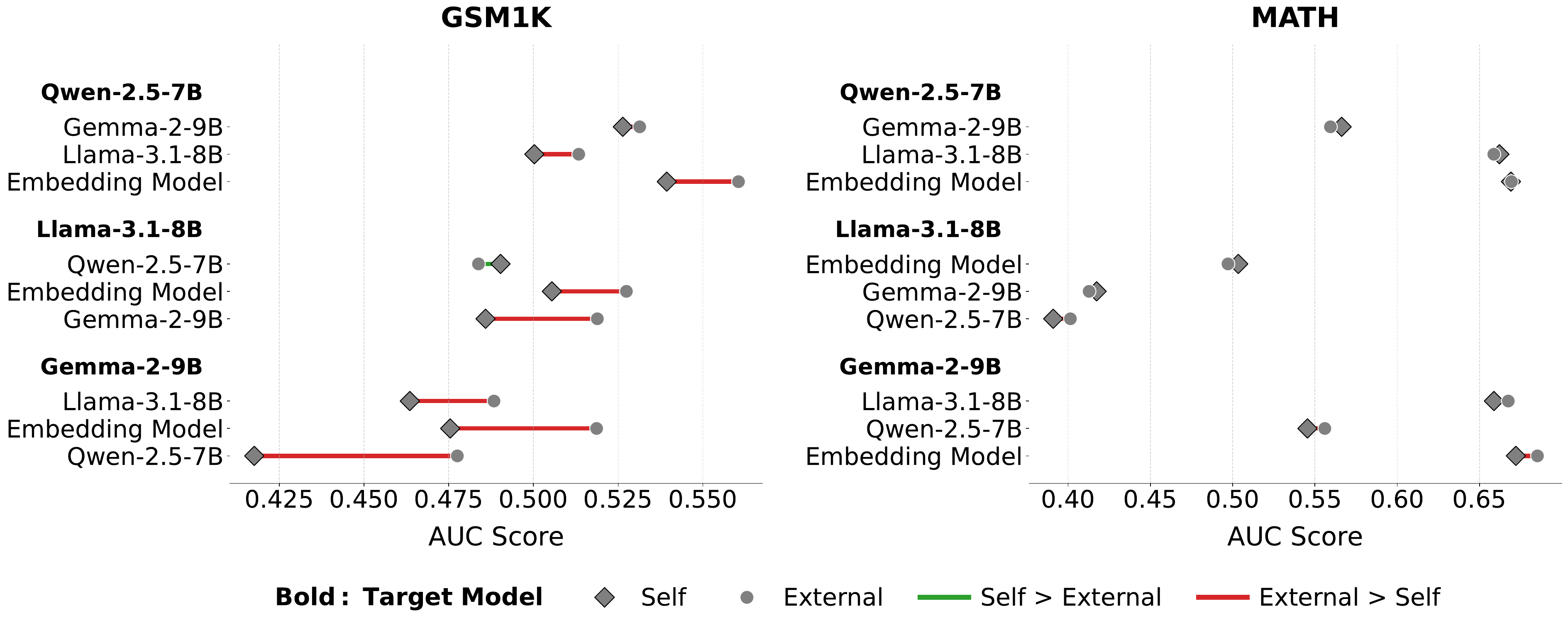}
    \caption{\textbf{Disagreement Gap: Mathematical Reasoning Breakdown (MLP Probes).}
    Detailed disagreement subset performance using MLP probes across GSM1K and MATH.
    Consistent with linear results, increased probe expressivity does not uncover hidden privileged info in math tasks; external models remain effective predictors.}
    \label{fig:disagreement_dumbbell_math_mlp}
\end{figure*}

Detailed results for factual knowledge tasks are shown in \Cref{fig:disagreement_dumbbell_factual_lr} (linear) and \Cref{fig:disagreement_dumbbell_factual_mlp} (MLP). Corresponding results for mathematical reasoning tasks are presented in \Cref{fig:disagreement_dumbbell_math_lr} (linear) and \Cref{fig:disagreement_dumbbell_math_mlp} (MLP).

\section{Implementation Details}
\label{app:implementation_details}

\subsection{Probe Training and Hyperparameters}
All probes were trained using a Stratified K-Fold Cross-Validation scheme with $k=10$ outer folds to estimate generalization performance. AUC metric is computed from pooled out-of-fold (OOF) predictions across these splits.

\textbf{Linear Probe:} We used Logistic Regression with $L_2$ regularization and the \texttt{liblinear} solver. Hyperparameters were selected using an inner $3$-fold cross-validation on the training split of each outer fold, tuning the regularization strength $C \in \{0.01, 0.1\}$. The model was trained with balanced class weights, standardized inputs, and a maximum of $500$ iterations.

\textbf{MLP Probe:} We used a Multi-Layer Perceptron (MLP) classifier with a single hidden layer of size $(100,)$ and ReLU activation. Hyperparameters were fixed across folds (i.e., no inner cross-validation), with an $L_2$ penalty of $\alpha=0.1$. The model was trained with early stopping enabled (using a $10\%$ validation split), standardized inputs, and a maximum of $500$ training iterations.

\subsection{Significance Testing}
To assess statistical uncertainty, we computed $95\%$ confidence intervals (CIs) using bootstrap resampling over the pooled out-of-fold (OOF) predictions. Specifically, we resampled the OOF predictions with replacement for $1000$ iterations and computed the empirical $2.5$ and $97.5$ percentiles of the resulting AUC distribution.

\section{Dataset Generation and Evaluation Details}
\label{app:dataset_details}

We standardized our generation and evaluation protocols using official model pipelines, aligning our methodology with the Language Model Evaluation Harness \citep{eval-harness} to ensure reproducibility.

\subsection{Response Generation}
Models were loaded using standard Hugging Face integration. We utilized greedy decoding (\texttt{do\_sample=False}) across all experiments to ensure deterministic outputs. Input prompts followed standard dataset-specific templates. Generation lengths were strictly controlled based on the domain: we set \texttt{max\_new\_tokens=32} for factual knowledge tasks to enforce concise entity generation, and \texttt{max\_new\_tokens=2048} for mathematical reasoning to accommodate full Chain-of-Thought derivations.

\subsection{Correctness Evaluation}
\paragraph{Factual Knowledge.} For Mintaka, TriviaQA, and HotpotQA, we evaluated correctness using standard Exact Match criteria. A response was labeled correct if any valid alias from the ground truth appeared as a case-insensitive substring within the generated text.

\paragraph{Mathematical Reasoning.}
For MATH and GSM1K, correctness was evaluated using the official evaluation scripts provided with each dataset. These scripts perform robust answer extraction by parsing the generated text to identify the final answer (e.g., via LaTeX \texttt{\textbackslash boxed\{\}} markers when present) and verify symbolic equivalence with the ground-truth solution, accounting for algebraic and notational variations.

\subsection{Distributional Shifts vs. \\ Privileged Knowledge}
\label{app:distributional_shift}

A potential alternative explanation for the premium gap observed 
on disagreement subsets is a general distributional shift near 
the decision boundary: harder examples may induce 
representational differences that benefit self-probes 
independently of any privileged knowledge. However, this 
explanation predicts that both self and peer probes should 
degrade similarly when restricted to disagreement subsets, as 
the increased difficulty would affect all probes equally. 
Instead, peer probes degrade substantially on disagreement 
subsets while self-probes retain a consistent and statistically 
significant advantage, suggesting the premium gap reflects 
genuine privileged signals specific to the target model's 
internal states rather than a general distributional artifact.

\section{Per-Layer Analysis: Additional Figures}
\label{app:layer_analysis_figures}

\begin{figure*}[t]
    \centering
    \begin{subfigure}[t]{\linewidth}
        \centering
        \includegraphics[width=\linewidth]{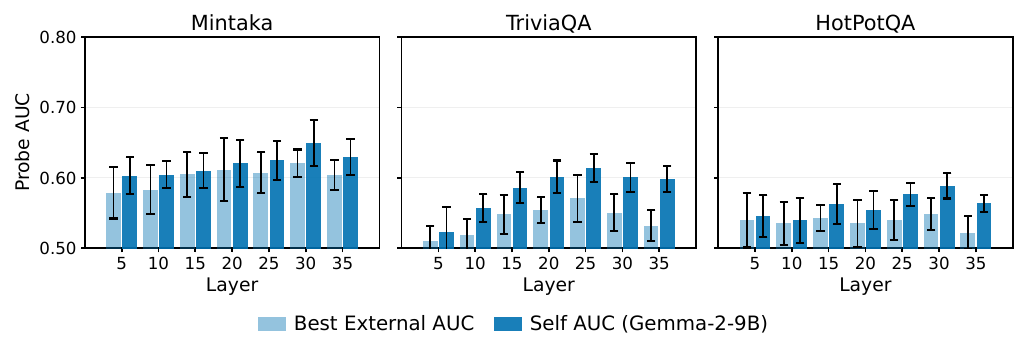}
        \caption{Factual Knowledge (Mintaka, TriviaQA, HotPotQA).}
        \label{fig:layer_bar_factual_gemma}
    \end{subfigure}
    \vspace{0.5em}
    \begin{subfigure}[t]{\linewidth}
        \centering
        \includegraphics[width=\linewidth]{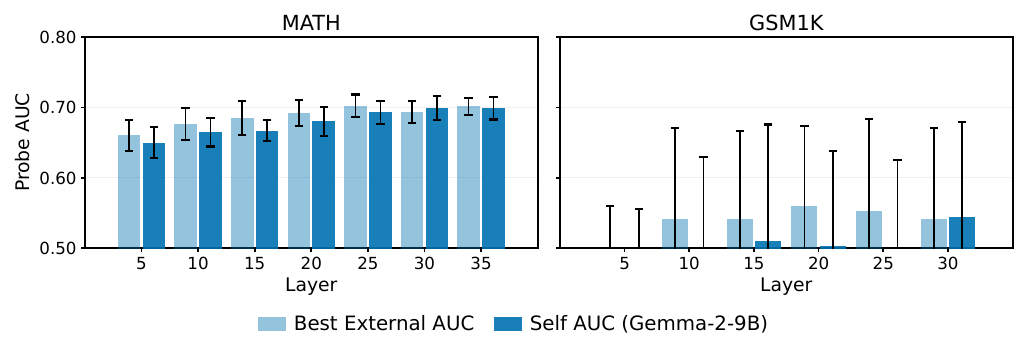}
        \caption{Mathematical Reasoning (MATH, GSM1K).}
        \label{fig:layer_bar_math_gemma}
    \end{subfigure}
    \caption{\textbf{Per-Layer AUC: Gemma-2-9B.}
    Absolute AUC at each probed layer on the disagreement subset. Lighter bars: best external probe; darker bars: self-probe.
    \textbf{(a)}~For factual datasets, the self-probe advantage grows visibly from mid-layers onward, particularly in TriviaQA.
    \textbf{(b)}~For mathematical reasoning, bars are of similar or reversed height, consistent with the absence of a premium gap. Error bars denote 95\% confidence intervals from cross-validation fold scores.}
    \label{fig:layer_bar_gemma}
\end{figure*}

\begin{figure*}[t]
    \centering
    \begin{subfigure}[t]{\linewidth}
        \centering
        \includegraphics[width=\linewidth]{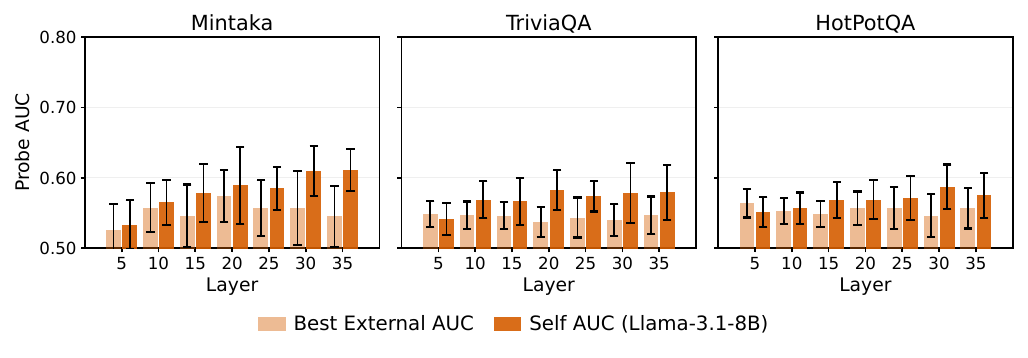}
        \caption{Factual Knowledge (Mintaka, TriviaQA, HotPotQA).}
        \label{fig:layer_bar_factual_llama}
    \end{subfigure}
    \vspace{0.5em}
    \begin{subfigure}[t]{\linewidth}
        \centering
        \includegraphics[width=\linewidth]{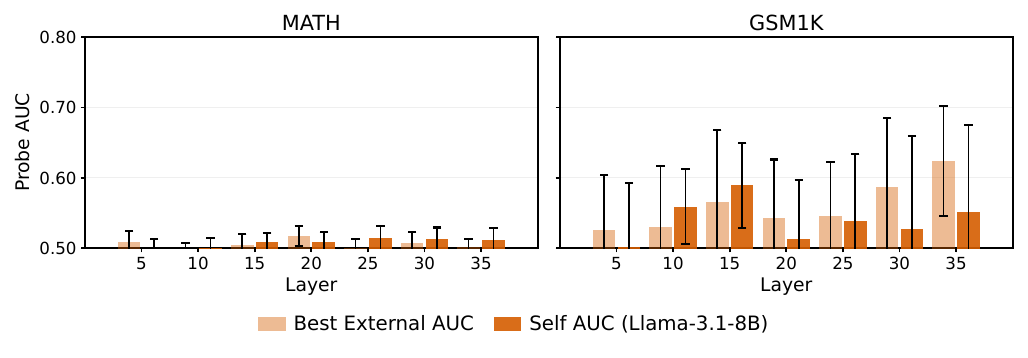}
        \caption{Mathematical Reasoning (MATH, GSM1K).}
        \label{fig:layer_bar_math_llama}
    \end{subfigure}
    \caption{\textbf{Per-Layer AUC: Llama-3.1-8B.}
    Absolute AUC at each probed layer on the disagreement subset. Lighter bars: best external probe; darker bars: self-probe.
    \textbf{(a)}~For factual datasets, the self-probe advantage emerges in early-to-mid layers.
    \textbf{(b)}~For mathematical reasoning, bars are of similar or reversed height, consistent with the absence of a premium gap. Error bars denote 95\% confidence intervals from cross-validation fold scores.}
    \label{fig:layer_bar_llama}
\end{figure*}

\begin{figure*}[t]
    \centering
    \begin{subfigure}[t]{\linewidth}
        \centering
        \includegraphics[width=\linewidth]{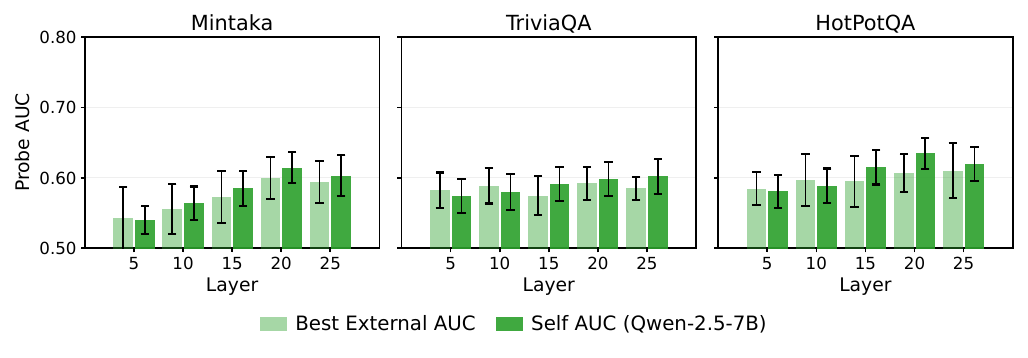}
        \caption{Factual Knowledge (Mintaka, TriviaQA, HotPotQA).}
        \label{fig:layer_bar_factual_qwen}
    \end{subfigure}
    \vspace{0.5em}
    \begin{subfigure}[t]{\linewidth}
        \centering
        \includegraphics[width=\linewidth]{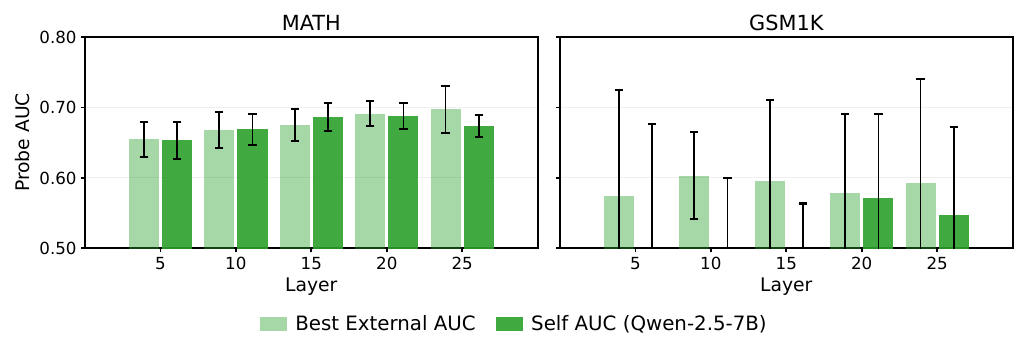}
        \caption{Mathematical Reasoning (MATH, GSM1K).}
        \label{fig:layer_bar_math_qwen}
    \end{subfigure}
    \caption{\textbf{Per-Layer AUC: Qwen-2.5-7B.}
    Absolute AUC at each probed layer on the disagreement subset. Lighter bars: best external probe; darker bars: self-probe.
    \textbf{(a)}~For factual datasets, the self-probe advantage emerges in early-to-mid layers.
    \textbf{(b)}~For mathematical reasoning, bars are of similar or reversed height, consistent with the absence of a premium gap. Error bars denote 95\% confidence intervals from cross-validation fold scores.}
    \label{fig:layer_bar_qwen}
\end{figure*}

\Cref{fig:layer_premium_gap_factual,fig:layer_premium_gap_math} in the main text shows the per-layer premium gap for both factual and mathematical reasoning datasets. Here we provide per-model bar figures showing absolute self and best-external AUC values at each probed layer on the disagreement subset (\Cref{fig:layer_bar_gemma,fig:layer_bar_llama,fig:layer_bar_qwen}). In each bar figure, the lighter bar represents the best external-probe AUC and the darker bar represents the self-probe AUC; the premium gap is directly visible as the height difference.

\section{Probing the Sources of Correctness Signals}
\label{app:lexical_only}

Our main analysis establishes that privileged knowledge emerges in factual tasks but not in mathematical reasoning. A related question is what information drives correctness prediction in each domain. To test whether entity identity alone can partially account for the correctness signal, we introduce a \textbf{Lexical-Only} control: we retain only named entities and nouns from the question (discarding syntax and function words) and train probes on the target model's hidden states when processing this stripped input. If correctness prediction is driven by entity-level familiarity, entity tokens alone should recover substantial performance even without syntactic context.

\paragraph{Implementation.}
We extract concepts from the original question using a two-stage pipeline: (1)~named entities via \texttt{GLiNER} (\texttt{urchade/gliner\_medium-v2.1}) across 20 label types, and (2)~noun chunks via \texttt{spaCy} (\texttt{en\_core\_web\_sm}). The union is deduplicated, removing substrings and stopwords. The resulting concepts are formatted as: \textit{``This text discusses [Concept~A], [Concept~B], and [Concept~C].''} This synthesized text replaces the original question as input to the target model, and probes are trained on the extracted hidden states to predict correctness on the \textit{original} question. All other experimental configurations follow \Cref{sec:exp_setup}.

\begin{figure*}[t]
    \centering
    \includegraphics[width=1.0\linewidth]{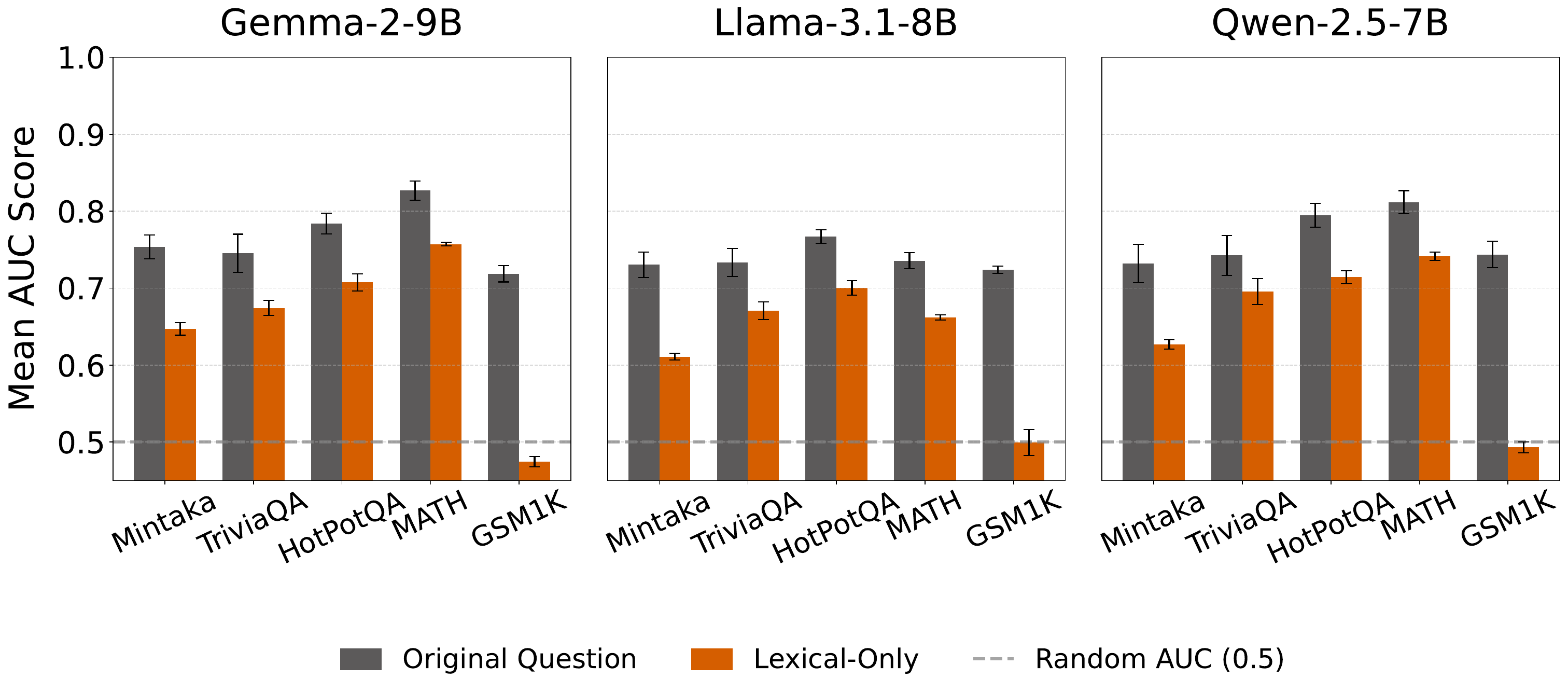}
    \caption{\textbf{Lexical-Only vs.\ Original Question.}
    Mean AUC for correctness prediction, averaged over layers, of probes trained on 
    the \textit{Original Question} versus the \textit{Lexical-Only} input (named 
    entities and nouns only), aggregated across all models (\texttt{Gemma-2-9B}, 
    \texttt{Llama-3.1-8B}, \texttt{Qwen-2.5-7B}). The gap between conditions 
    reflects the contribution of syntactic and contextual processing beyond entity 
    identity. Error bars denote 95\% confidence intervals from cross-validation fold scores.}
    \label{fig:bags_aggregated_all}
\end{figure*}

\paragraph{Results.}
As shown in \Cref{fig:bags_aggregated_all}, Lexical-Only probes recover a non-trivial portion of the original probe's performance across all datasets except GSM1K. In factual domains (Mintaka, TriviaQA, HotpotQA), lexical features recover $53.7\%$, $75.0\%$, and $73.5\%$ of original probe performance relative to the random baseline ($0.5$ AUC), suggesting that much of the correctness signal stems from concept-level familiarity. MATH shows a similar pattern, recovering $75.6\%$ of performance, likely because lexical tokens encode mathematical topic indicators (e.g., ``eigenvalue'', ``asymptote'') that correlate with difficulty. In contrast, GSM1K drops to near chance (AUC $\approx 0.49$), consistent with its correctness being governed by computational structure rather than surface tokens such as ``savings account'' or ``\$50''. These results indicate that correctness prediction in factual tasks, and to some extent in MATH, relies substantially on concept-level signals, whereas GSM1K correctness depends on structural problem features that lexical stripping destroys.

\section{Hardware Details}
\label{app:hardware}
All experiments were conducted on a system with 32 Intel(R) Xeon(R) Gold 6430 CPUs and 1.0~TB of RAM. The system was equipped with three NVIDIA RTX 6000 Ada Generation GPUs, each with 49~GB of VRAM.

\section{Licenses and Third-Party Usage}
\label{app:licenses}
This work is implemented using \textbf{PyTorch} \citep{paszke2019pytorch}, an open-source deep learning framework licensed under the BSD license, and the \textbf{Hugging Face Transformers} library \citep{wolf2019huggingface}, licensed under Apache 2.0. We also employ \textbf{spaCy} (MIT License) and \textbf{GLiNER} (Apache 2.0) for the lexical analysis described in the control experiments. All software usage complies with their respective license terms.

\paragraph{Datasets.} We utilize several open-source datasets for evaluation:
\begin{itemize}
    \item \textbf{Mintaka} \citep{sen2022mintaka} is licensed under CC-BY 4.0.
    \item \textbf{HotpotQA} \citep{yang2018hotpotqa} is licensed under CC-BY-SA 4.0.
    \item \textbf{TriviaQA} \citep{joshi2017triviaqa} is licensed under Apache 2.0.
    \item \textbf{GSM1K} \citep{zhang2024careful} and \textbf{MATH} \citep{hendrycks2measuring} are licensed under the MIT License.
\end{itemize}

\section{Use of AI Assistants}
\label{app:ai_assistants}
We utilized AI assistants for refining text clarity and coding assistance. All scientific claims, experimental results, and final text were written by the authors.

\end{document}